\documentclass[acmtog]{acmart}
\acmSubmissionID{168}

\usepackage[boxed,ruled]{algorithm2e} 

\SetCommentSty{mycommfont}
\let\oldnl\nl
\newcommand{\nonl}{\renewcommand{\nl}{\let\nl\oldnl}}

\SetAlFnt{\small}
\SetAlCapFnt{\small}
\SetAlCapNameFnt{\small}
\SetAlCapHSkip{0pt}
\IncMargin{-\parindent}

\newlength\savedwidth
\newcommand\whline[1]{\noalign{\global\savedwidth\arrayrulewidth
                               \global\arrayrulewidth #1} %
                      \hline
                      \noalign{\global\arrayrulewidth\savedwidth}}

\usepackage{booktabs} 
\usepackage{enumitem}
\usepackage{amsmath}
\usepackage{mathtools}

\usepackage{color, colortbl}
\definecolor{lightgray}{gray}{0.9}
\usepackage{xcolor}
\usepackage{wrapfig}
\usepackage{eucal}
\DeclareMathAlphabet\mathbfcal{OMS}{cmsy}{b}{n}

\newcommand{\blue}[1]{\textcolor{blue}{#1}}

\begin{document}
\title{High-order Differentiable Autoencoder for Nonlinear Model Reduction}


\author{Siyuan Shen}
\affiliation{%
 \institution{State Key Lab of CAD\&CG, Zhejiang University}
 \streetaddress{866 Yuhangtang Rd}
 \city{Hangzhou}
 \postcode{310058}
 \country{China}}
\email{shensiyuan@zju.edu.cn}

\author{Yang Yin}
\affiliation{%
 \institution{Clemson University}
 \country{USA}}
\email{yin5@clemson.edu}

\author{Tianjia Shao}
\affiliation{%
 \institution{State Key Lab of CAD\&CG, Zhejiang University}
 \country{China}}
\email{tianjiashao@gmail.com}

\author{He Wang}
\affiliation{%
 \institution{University of Leeds}
 \country{United Kingdom}}
\email{H.E.Wang@leeds.ac.uk}

\author{Chenfanfu Jiang}
\affiliation{%
 \institution{University of Pennsylvania}
 \country{USA}}
\email{cffjiang@seas.upenn.edu}

\author{Lei Lan}
\affiliation{%
 \institution{Clemson University}
 \country{USA}}
\email{lanlei.virhum@gmail.com}

\author{Kun Zhou}
\affiliation{%
 \institution{State Key Lab of CAD\&CG, Zhejiang University}
 \country{China}}
\email{kunzhou@acm.org}

\renewcommand\shortauthors{Shen, S. et al}

\begin{abstract}
This paper provides a new avenue for exploiting deep neural networks to improve physics-based simulation. Specifically, we integrate the classic Lagrangian mechanics with a deep autoencoder to accelerate elastic simulation of deformable solids. Due to the inertia effect, the dynamic equilibrium cannot be established without evaluating the second-order derivatives of the deep autoencoder network. This is beyond the capability of off-the-shelf automatic differentiation packages and algorithms, which mainly focus on the gradient evaluation. Solving the nonlinear force equilibrium is even more challenging if the standard Newton's method is to be used. This is because we need to compute a third-order derivative of the network to obtain the variational Hessian. We attack those difficulties by exploiting complex-step finite difference, coupled with reverse automatic differentiation. This strategy allows us to enjoy the convenience and accuracy of complex-step finite difference and in the meantime, to deploy complex-value perturbations as collectively as possible to save excessive network passes. With a GPU-based implementation, we are able to wield deep autoencoders (e.g., $10+$ layers) with a relatively high-dimension latent space in real-time. Along this pipeline, we also design a sampling network and a weighting network to enable \emph{weight-varying} Cubature integration in order to incorporate nonlinearity in the model reduction. We believe this work will inspire and benefit future research efforts in nonlinearly reduced physical simulation problems.
\end{abstract}

%
%

\ccsdesc[500]{Computing methodologies~Physical simulation}
\ccsdesc[500]{Computing methodologies~Dimensionality reduction and manifold learning}

%
%

\keywords{Model reduction, Autoencoder, Differentiation, GPU, Deformable model}

\begin{teaserfigure}
\center
  \includegraphics[width=\textwidth]{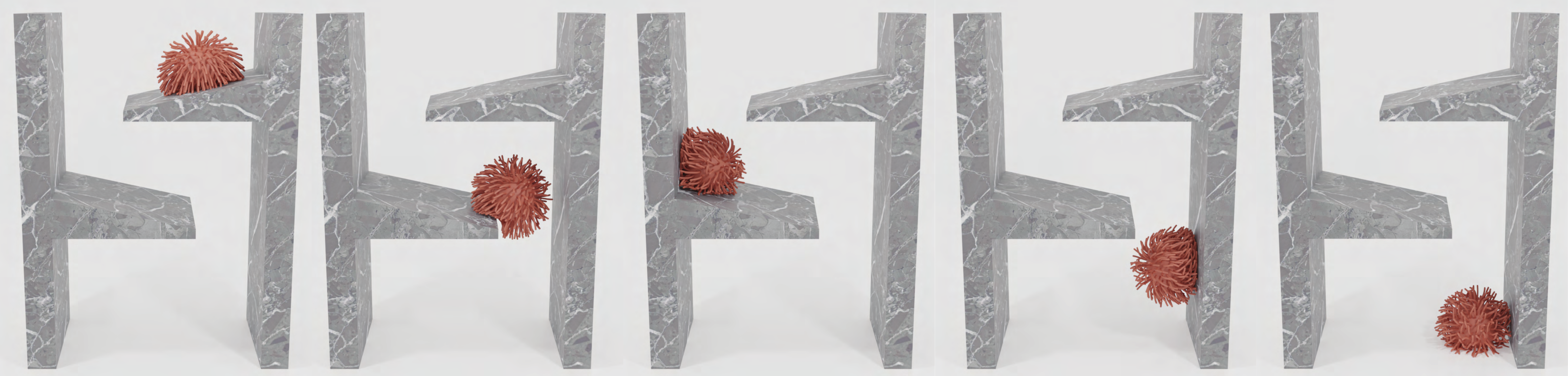}
  \caption{In this paper, we exploit deep autoencoder (DAE) networks to accelerate physics-based simulation. In order to model nonlinear subspace dynamics accurately, second- and high-order derivatives of the deep decoder net must be efficiently evaluated to match the subspace simulation frame rate. We address this technical challenge by collectively applying complex-step network perturbations to the deep net. This is the first time a high-order differentiable neural net is employed in physical simulation problems. Our method can be further strengthened with the domain decomposition method as a nonlinear DAE better captures local deformation effects. In this example, the puffer ball has $320$ elastic strings, and we assign a $n_p=10$ linear subspace and a $n_q=5$ nonlinear subspace at each string. With the help of substructured deformation, DAE-based nonlinear reduction produces interesting animation effects. We believe this is a representative example showing case the advantage of data-driven animation using DAE. As the geometries of all the strings are the same, generating local training poses is more effective.}
  \label{fig:teaser}
\end{teaserfigure}
\maketitle

\section{Introduction}\label{sec:intro}

Model reduction is a widely-used and highly-effective technique for accelerating physically-based simulation. It is also sometimes known as reduced-order simulation or subspace simulation. While named differently, the core idea is to build a linear subspace with reduced degrees-of-freedom (DOFs) so that the physical equations can be solved with a system of a smaller size. This approach is sensible because many parts of the real physical world evolve smoothly and continuously along the time and space. Sharp and high-frequency physical changes are less common and should be treated with dedicated numerical methods. Existing model reduction methods have been dominantly \emph{linear} reduction with a constant tangent space. The expressivity of linear reduction is a known limitation. As many physical phenomena are intrinsically nonlinear, a linearly reduced model only covers a small fraction of the dynamics space -- all the information outside of the subspace is filtered. Thus, one has to (substantially) increase the dimensionality of the subspace to incorporate a desired nonlinear effect even this effect itself may be of low rank (thinking of a bead travelling on the circle).

One question rises naturally: can we build a nonlinear reduction framework with a time-varying (as opposed to constant) tangent space that best fits ``local'' dynamics? The challenges are twofold. First, the underlying manifold representing the nonlinear dynamics is often too complex to be expressed in a closed form. Developing a nonlinear subspace-fullspace transformation, as a counterpart of modal analysis in linear reduction, is theoretically difficult. Second, a nonlinear reduction brings extra computation burdens to the simulation, largely originating from the need for evaluating the derivatives of the subspace-fullspace transformation function. The computational cost goes up quickly with respect to the subspace size, which in turn neutralizes the original motivation of applying model reduction.

In this paper, we propose a new nonlinear model reduction framework that is tightly coupled with the classic Lagrangian mechanics. Although we demonstrate the effectiveness of our method in the context of elastic simulation, we believe our method could also be useful in other physics-based simulation problems like fluid~\cite{kim2013subspace} or cloth animations~\cite{hahn2014subspace}. Our reduction mechanism is data-driven with a deep autoencoder (DAE) in the loop, which obviates the need for a closed-form subspace-fullspace transformation function. DAE is an unsupervised learning architecture skilled in data compression~\cite{hinton2006reducing} and has been proven effective in deformable simulation recently~\cite{fulton2019latent}. Along this direction, we augment the DAE network with the complex-step finite difference (CSFD) method~\cite{martins2003complex}, enabling its high-order differentiability, so that the neural net can be computationally integrated with Lagrangian formulation. To achieve this goal, we make several mentionable technical contributions:

\vspace{5 pt}
\noindent$\bullet$~~\textit{Efficient high-order differentiable deep autoencoder.}\\
A physically accurate coupling between DAE and elastic dynamics requires the information of the first- and second-order derivatives (for Newtonian equations of motion). Existing differentiation techniques such as backpropagation (BP)~\cite{hecht1992theory} for neural networks or automatic differentiation (AD)~\cite{bucker2006automatic} for more general computations are optimized for gradient estimation only, and become cumbersome in high-order cases. We leverage CSFD to facilitate the differentiation of the encoder network. Conceptually straightforward, this however is not ``as easy as pie'' as it appears. Albeit the excellent numerical robustness and accuracy, CSFD needs to apply a complex-value perturbation for each input variable, leading to excessive forward passes of the deep neural net. We resolve this challenge by applying the function perturbation \emph{collectively} and deploying CSFD \emph{inside} other differentiation procedures such as BP and directional derivative. With a GPU-based implementation, we simulate complex nonlinear models in real time with a deep decoder net in the loop.

\vspace{5 pt}
\noindent$\bullet$~~\textit{Coupling PCA with deep encoding net.}\\
The tangent space of a DAE  may vary drastically to incorporate nonlinearity seen in the training poses yielding a bumpy and uneven deformation manifold. This is analogous to over-fitting. A possible cure is to use contractive autoencoder or CAE~\cite{rifai2011contractive}. CAE adds a regularizer in the objective function that forces the network to learn a function that is robust under slight input variations. While it could be a viable solution, we propose a more convenient and effective option. In our framework, DAE is constructed within the residual space of a standard PCA. In other words, DAE is designed to be complementary to an underlying linear subspace, and the latter guarantees the existence of a smooth tangent variation. With this design, we can deepen the DAE architecture to capture nonlinear and salient deformation poses.

\vspace{5 pt}
\noindent$\bullet$~~\textit{Weight-varying subspace integration using deep neural networks.}\\
Reduced simulation is often coupled with sparse force and Hessian integration, which down samples element-wise volumetric integration to a small collection of key elements, called Cubature elements~\cite{an2008optimizing}. After Cubature elements are selected, one also needs to compute its integration weight by solving a non-negative least-square problem. The weight coefficient of a Cubature element is typically fixed given the training set. This is reasonable for linear reduction and works well in practice. However, in the context of nonlinear reduction, as the tangent space varies along the simulation, fixed weighting Cubature is problematic. To this end, we propose a deep neural network (DNN) based sampling method, that fully replaces Cubature training. Our DNN has two modules. The first module is a graph convolution network (GCN) that outputs the possibility of an element being a Cubature element. On the top of it, the second module is a DNN, which predicts the weight of selected Cubature elements. The last layer of this DNN carries out a per-neuron square operation to ensure the final network output is non-negative. The training alternates between those two modules. Unlike conventional Cubature sampling strategy, our network-based approach is able to select multiple Cubature elements each iteration, thus greatly shortens the training time.

We have evaluated our framework on various simulation scenarios, and our method produces visually-plausible results in real time or at an interactive rate. We also notice that our nonlinear model reduction framework synergizes with domain decomposition method~\cite{barbivc2011real,yang2013boundary,wu2015unified} -- a small-size nonlinear subspace captures deformation effects much better at a local domain than over the entire deformable body. To this end, we also demonstrate examples combining DAE and domain decomposition. As a natural follow up of our method, we do not intend to over claim this extension as our contribution. Model reduction, regardless nonlinear or linear, seeks for smart trade-offs among simulation effects, accuracy, and performance. Arguing conclusively that the nonlinear model reduction is always better than linear model reduction techniques is too bold and over-confident to us. Indeed, one should scrutinize various aspects in practice, such as the problem size, expected results, time budget, hardware resources etc. before choosing a specific simulation algorithm. Regardless, we do believe the techniques purposed in the paper are worthy and non-trivially advance state-of-the-art model reduction methods.

\section{Related Work}\label{sec:related}
Model reduction has been successfully employed in many simulation-related problems in computer graphics including fluid dynamics~\cite{treuille2006model,kim2013subspace}, cloth animation~\cite{hahn2014subspace}, shape deformation~\cite{von2015real,wang2015linear}, material design~\cite{xu2015interactive,musialski2016non}, animation control~\cite{barbivc2009deformable,barbivc2012interactive} etc. In this paper, we narrow our focus on using data-driven nonlinear model reduction to improve elastic simulation of solid objects.

There are several well-established numerical solutions for deformable models such as finite element method (FEM)~\cite{zienkiewicz1977finite}, finite difference method~\cite{zhu2010efficient}, meshless method~\cite{martin2010unified,muller2005meshless}, or mass-spring system~\cite{Liu2013}. Most of them end up with solving a large-scale nonlinear system if an implicit time integration scheme is used. For high-resolution models, computing their time-varying nonlinear dynamics is expensive. Speeding up the deformable simulation can be achieved using carefully designed numerical treatments like the multigrid method~\cite{zhu2010efficient,tamstorf2015smoothed}, delayed matrix update~\cite{hecht2012updated}, or parallelizable solvers~\cite{wang2016descent,fratarcangeli2016vivace}. These methods focus on improving the performance for the \emph{fullspace} nonlinear optimization without condensing the simulation scale.

Acceleration can also be achieved using model reduction, which removes less important DOFs and creates a subspace representation of fullspace DOFs. Modal analysis~\cite{pentland1989good,hauser2003interactive,choi2005modal} and its first-order derivatives~\cite{barbivc2005real,yang2015expediting} 
are often considered as the most effective way for the subspace construction. Displacement vectors from recent fullspace simulations can also be utilized as subspace bases~\cite{kim2009skipping}. Alternatively, it is also viable to coarsen geometric representation to prescribe the dynamics of a fine model like skin rigging, a technique widely used in animation systems~\cite{james2005skinning}. Analogously, Capell and colleagues~\shortcite{capell2002interactive} deformed an elastic body using an embedded skeleton; Gilles and colleagues~\shortcite{gilles2011frame} used rigid frames to drive the deformable simulation; Faure and colleagues~\shortcite{faure2011sparse} used scattered \emph{handles} for reduced models; Lei and colleagues~\shortcite{lan2020medial} combined model reduction with collision processing using medial axis transform; Martin and colleagues~\shortcite{martin2010unified} used sparsely-distributed integrators named \emph{elastons} to model the nonlinear dynamics of rod, shell, and solid uniformly.

Those prior arts demonstrate impressive results, often with orders-of-magnitude performance speedups with model reduction. In most cases, the generalized coordinate linearly depends on the fullspace displacement in the form of $\mathbf{u} = \mathbf{U}\mathbf{q}$ with $\mathbf{U}$ being constant. This is why we refer to them as linear subspace methods. On the contrary, nonlinear reduction holds a more complicated relation between generalized and fullspace coordinates. For instance, nonlinear modal analysis~\cite{pesheck2001nonlinear} aims to extend its linear version, and it has been used in structural analysis~\cite{setio1992modal}. However, it is hardly useful for simulation acceleration -- extracting the modal space for a given configuration is normally dealt with by solving an eigenproblem (i.e., as in linear modal analysis), and it is clearly infeasible to exhaustively sample all the system configurations even in the pre-computation stage. Only when the animation follows some pre-known patterns, we may re-use the solution of rest-shape eigenproblem~\cite{mukherjee2016incremental} or interpolate multiple sparsely chosen linear subspaces~\cite{xu2016pose}. Due to this challenge, the nonlinear subspace method is less explored.

In this paper, we do not aim to derive a closed-form mathematical formulation connecting the generalized coordinate and the fullspace coordinate. Instead, we leave this challenge to a deep neural network that learns the map directly from many seen simulation poses. This is a straightforward data-driven approach and has been exploited in graphics for years~\cite{wang2011data,ladicky2015data}. Our novelty however is to \emph{enable its efficient and high-order differentiability so that the DNN can be embedded into the classic physical simulation frameworks} such as Lagrangian mechanics~\cite{brizard2014introduction} etc. The DNN used in our framework is a deep autoencoder or DAE~\cite{hinton2006reducing} originally designed for dimension reduction. Its superior performance in data compression and multidimensional scaling has quickly drawn many attentions. DAE is successfully deployed in NLP~\cite{socher2011semi}, image/video compression~\cite{balle2016end,habibian2019video}, GAN~\cite{makhzani2015adversarial}, facial recognition~\cite{zeng2018facial}, 3D shape analysis~\cite{nair20093d}, just to name a few. The volume of DAE-related studies is too vast to be contained here. 

The most relevant study of our work is the contribution from Fulton and colleagues~\shortcite{fulton2019latent}. Indeed, we are strongly motivated and inspired by those recent efforts~\cite{wiewel2019latent,fulton2019latent} that also seek for DAE-based nonlinear reduction. To this end, we re-examine each step along the pipeline of reduced simulation and devise a comprehensive solution to couple DAE with nonlinear elastic simulation seamlessly. One core ingredient is the high-order differentiability that should be evaluated efficiently to match the frame rate of model reduction. We deliver this important technical asset by leveraging complex-step finite difference or CSFD~\cite{martins2003complex}. Similar to standard finite difference, CSFD applies a perturbation to the function input and evaluates how the perturbation alters the function output. This perturbation however, is a complex-value quantity in CSFD, which avoids the numerical issue of subtractive cancellation. While CSFD seems to be a possible approach to the differentiability of DAE, its efficient deployment for high-order differentiation remains challenging in practice. An na\"ive implementation of CSFD requires one forward pass of the net for each input variable. This scheme leads to $\mathbf{O}(n^3)$ passes for a DAE-enabled Newton iteration. As discussed in \S~\ref{sec:CSFD}, we attack this difficulty by applying the complex stepping in CSFD collectively whenever possible to remove redundant network passes. This method allows us to efficiently run the DAE differentiation on GPU and obtain its high-order derivatives in milliseconds. In order to ensure the smoothness of the simulation tangent space, our framework consists of two layers: the first layer is a standard PCA-based linear subspace and within the orthogonal space of which, DAE is deployed to capture nonlinear deformations more effectively. Lastly, we also design a DNN-based Cubature training procedure to generate pose-dependent weight coefficients for a more accurate subspace integration.

\section{Linear and Nonlinear Model Reduction}\label{sec:reduction}
To make the paper more self-contained, we start with a brief review of the linear model reduction framework, and show its nonlinear generalization with DAE afterwards. Here, we assume that a DAE is differentiable and defer the discussion about how to compute its first- and high-order derivatives to the next section. 

\subsection{Linear Model Reduction}
Under the FEM discretization, the motion of an elastically deformable solid can be described with the Euler-Lagrange equation:
\begin{equation}\label{eq:equation_of_motion}
    \mathbf{M}\ddot{\mathbf{u}} + \mathbf{f}_{damp}(\mathbf{u}, \dot{\mathbf{u}}) + \mathbf{f}_{int}(\mathbf{u}) = \mathbf{f}_{ext},
\end{equation}
where $\mathbf{M}\in\mathbb{R}^{N \times N}$ is the fullspace mass matrix; $\mathbf{f}_{int}$ and $\mathbf{f}_{ext}$ are the nonlinear internal force and external force. Here, we lump $\mathbf{M}$ to be a diagonal matrix. $\mathbf{f}_{damp}$ is the damping force, and it is often modeled, under the assumption of Rayleigh damping, as:
\begin{equation}
\mathbf{f}_{damp} = \left(\alpha \mathbf{M} + \beta \frac{\partial \mathbf{f}_{int} (\mathbf{u})}{\partial \mathbf{u}}\right)\dot{\mathbf{u}}.
\end{equation}
Eq.~\eqref{eq:equation_of_motion} describes the force equilibrium at all $N$ DOFs of the unknown displacement vector $\mathbf{u}$. Computing $\mathbf{u}$ in  Eq.~\eqref{eq:equation_of_motion} using nonlinear methods like Newton's method needs to solve an $N$-by-$N$ linearized system repeatedly, which is slow and expensive for large-scale models. Linear model reduction prescribes the kinematic of this $N$-dimension system with a set of generalized coordinate $\mathbf{p}$ such that $\mathbf{u}=\mathbf{U}\mathbf{p}$. $\mathbf{U}\in\mathbb{R}^{N \times n}$ is sometimes called subspace matrix, which is constant in linear reduction. An important convenience brought by the linearity is the time derivatives of $\dot{\mathbf{u}} = \mathbf{U}\dot{\mathbf{p}}$ and $\ddot{\mathbf{u}} = \mathbf{U}\ddot{\mathbf{p}}$ follow the same relation. Therefore, Eq.~\eqref{eq:equation_of_motion} can be projected into the column space of $\mathbf{U}$ as:
\begin{equation}\label{eq:linear_reduction}
    \mathbf{M}_p\ddot{\mathbf{p}}+\mathbf{U}^\top\mathbf{f}_{damp}(\mathbf{U}\mathbf{p}, \mathbf{U}\dot{\mathbf{p}}) + \mathbf{U}^\top\mathbf{f}_{int}(\mathbf{U}\mathbf{p}) = \mathbf{U}^\top\mathbf{f}_{ext},
\end{equation}
where $\mathbf{M}_p=\mathbf{U}^\top\mathbf{M}\mathbf{U}$ is the reduced mass matrix. Eq.~\eqref{eq:linear_reduction} has the same structure of Eq.~\eqref{eq:equation_of_motion} despite under a more compact representation of $\mathbf{p}$.

\subsection{Nonlinear Model Reduction}
Nonlinear reduction also uses a set of generalized coordinates $\mathbf{q}$. However, the relation between $\mathbf{u}$ and $\mathbf{q}$ is in a more generic form of $\mathbf{u} = D(\mathbf{q})$. Intuitively, $D$ prescribes an $n$-dimension \emph{deformation manifold} embedded in $\mathbb{R}^N$. Applying time differentiation at both side yields: 
\begin{equation}\label{eq:velocity}
    \dot{\mathbf{u}} = \frac{\mathrm{d} D(\mathbf{q})}{\mathrm{d} t} = \frac{\partial D(\mathbf{q})}{\partial \mathbf{q}} \dot{\mathbf{q}} = \mathbf{J}\dot{\mathbf{q}},
\end{equation}
and 
\begin{equation}\label{eq:acceleration}
\ddot{\mathbf{u}}=\frac{\mathrm{d}}{\mathrm{d} t}\left( \frac{\partial D(\mathbf{q})}{\partial \mathbf{q}}  \dot{\mathbf{q}} \right) = (\mathcal{H}\cdot\dot{\mathbf{q}})\dot{\mathbf{q}}+\mathbf{J}\ddot{\mathbf{q}},
\end{equation}
where $\mathbf{J} = {\partial D(\mathbf{q})}/{\partial \mathbf{q}} \in \mathbb{R}^{N \times n}$ is the Jacobian of $D(\mathbf{q})$, which depends on $\mathbf{q}$ and spans the \emph{tangent space} at a given reduced coordinate. $\mathcal{H}\in\mathbb{R}^{N \times n \times n}$ is a third tensor of the second-order derivative (i.e., Hessian) of $D$. Substituting Eqs.~\eqref{eq:velocity}~and~\eqref{eq:acceleration} into Eq.~\eqref{eq:equation_of_motion} followed by a tangent space projection gives the nonlinearly reduced equation of motion:
\begin{equation}\label{eq:nonlinear_reduction}
\mathbf{J}^\top\mathbf{M}\big((\mathcal{H}\cdot\dot{\mathbf{q}})\dot{\mathbf{q}}+\mathbf{J}\ddot{\mathbf{q}}\big) + \mathbf{J}^\top\mathbf{f}_{int}(D(\mathbf{q})) = \mathbf{J}^\top\mathbf{f}_{ext}.
\end{equation}
Here, the damping force term is ignored for a more concise notation. We note that if $D(\mathbf{q})$ is linear, $\mathcal{H}$ vanishes and $\mathbf{J} = \mathbf{U}$. Eq.~\eqref{eq:nonlinear_reduction} echoes Eq.~\eqref{eq:linear_reduction} completely.

Given a time integration algorithm on $\mathbf{q}$ e.g., the implicit Euler method, we have: $\mathbf{q} = \bar{\mathbf{q}}  + h\dot{\mathbf{q}}$ and $\dot{\mathbf{q}} = \bar{\dot{\mathbf{q}}}  + h\ddot{\mathbf{q}}$, where $h$ is the time step size, and $\bar{(\cdot)}$ indicates the kinematic variable is from the previous time step. The final system that needs to be solved becomes:
\begin{equation}\label{eq:nonlinear_equilibrium}
    \mathbf{J}^\top\mathbf{M}\mathbf{J}(\mathbf{q}-\bar{\mathbf{q}} - h \bar{\dot{\mathbf{q}}}) + \mathbf{J}^\top\mathbf{f}_{fict}(\mathbf{q})
    + h^2\mathbf{J}^\top\mathbf{f}_{int}(\mathbf{q}) = h^2\mathbf{J}^\top\mathbf{f}_{ext},
\end{equation}
with
\begin{equation}\label{eq:fict}
    \mathbf{f}_{fict} = \mathbf{M}\left(\left[\mathcal{H}\cdot(\mathbf{q}-\bar{\mathbf{q}})\right](\mathbf{q}-\bar{\mathbf{q}})\right).
\end{equation}
$\mathbf{f}_{fict}$ is the fictitious force that responds for inertia effects associated with the varying Jacobian $\mathbf{J}$. We also consider $\mathbf{M}_q = \mathbf{J}^\top\mathbf{M}\mathbf{J}$ is the reduced mass matrix of the nonlinear reduction, which is no longer constant as $\mathbf{J}$ also depends on $\mathbf{q}$.

\subsection{A Quick Discussion}
Clearly, $\mathbf{f}_{fict}$ is the most tricky part in Eq.~\eqref{eq:nonlinear_equilibrium}. $\mathcal{H}$ is the Hessian of the coordinate transformation $D$. Not only a third tensor, but $\mathcal{H}(\mathbf{q})$ is also a function of $\mathbf{q}$. Therefore, if we want to solve Eq.~\eqref{eq:nonlinear_equilibrium} using, for instance, Newton's method in the implicit integration, we need to compute $\partial \mathcal{H}/\partial \mathbf{q}$ to assemble the corresponding system matrix, which is a forth tensor and the resultant of third-order differentiation over $D$. This nasty computation stands as a major obstacle for nonlinear model reduction. In~\cite{fulton2019latent}, the fictitious force term is discarded in the time integration of the generalized coordinate. This heuristic can somewhat be understood as performing an \emph{explicit} subspace projection at the current time step ignoring the fact that a generalized velocity $\dot{\mathbf{q}}$ also brings inertia effects when $D$ is nonlinear.

On the one hand, we consider ignoring $\mathbf{f}_{fict}$ reasonable and an understandable compromise in the setting of~\cite{fulton2019latent}. First of all, $\mathbf{f}_{fict}$ vanishes under quasi-static deformations as $\dot{\mathbf{q}}$ is close to zero. Secondly, in~\cite{fulton2019latent} the encoding-decoding network is shallow, and $D$ represents a net of only two layers. In addition, a preliminary PCA is performed to ``regularize'' raw training poses. Those treatments effectively suppress the nonlinearity in $\mathbf{J}$ (so $\mathcal{H}$ is small) and lessen the inertia deformation induced by $\mathbf{f}_{fict}$. On the other hand, as $\mathbf{f}_{fict}$ is missed in~\cite{fulton2019latent}, the underlying dynamic equation is inaccurate anyway. Visible artifacts are inevitable under higher-velocity deformations or a deeper DAE is employed (i.e., in our case).

\subsection{PCA-orthogonal DAE Reduction}
An autoencoder is an unsupervised learning algorithm that condenses the input high-dimension data into a low-dimension \emph{latent space} (i.e., encoding), which is then expanded to the original dimensionality to monitor the compression loss (i.e., decoding). If this network only has one hidden layer, or it does not involve nonlinear activations, the autoencoder is similar to PCA~\cite{bourlard1988auto}. In this case $D$ is linear, and the resulting network after training spans the same linear subspace as PCA does (under L2 loss).

\begin{wrapfigure}{r}{0.4\linewidth}
\vspace{-10 pt}
\includegraphics[width = \linewidth]{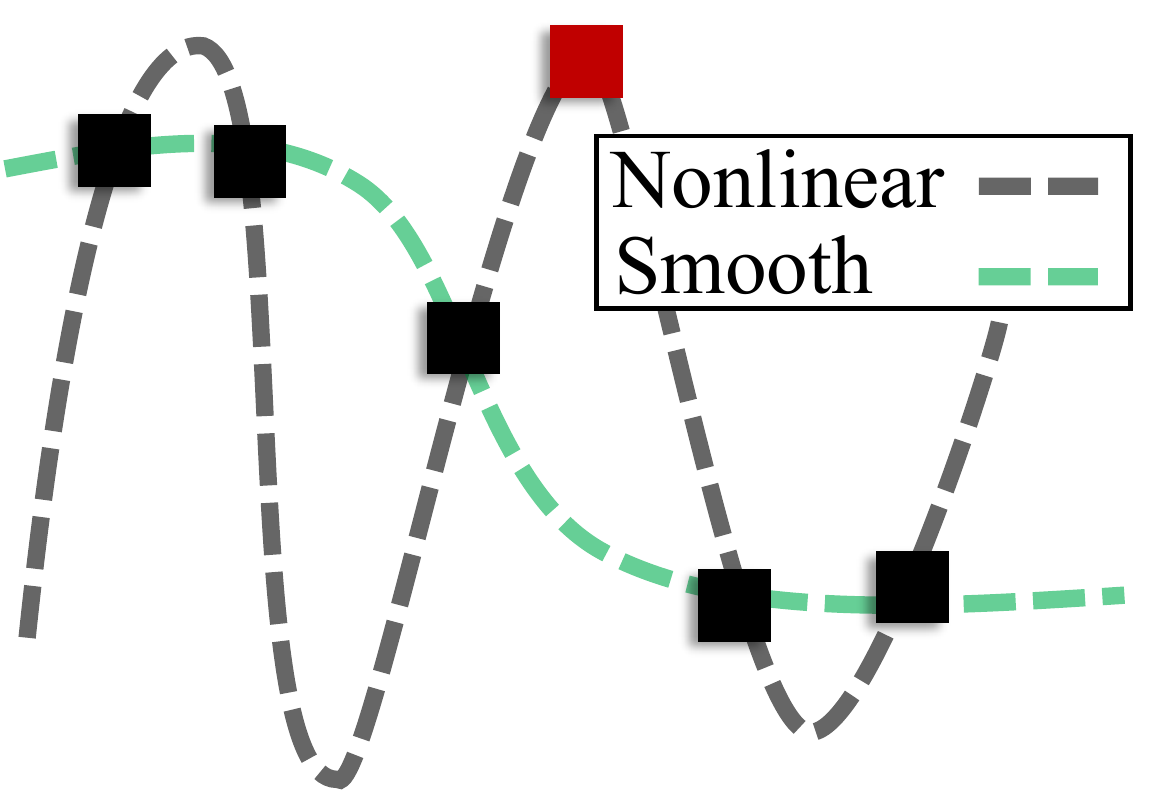}
\caption{Increased nonlinearity better fits training poses but also makes the network more wiggling.}
\vspace{-5 pt}
\label{fig:overfit}
\vspace{-5 pt}
\end{wrapfigure}

This however is, not what we seek for in nonlinear model reduction since $D$ is expected to capture as much nonlinearity as possible to enrich the subspace expressivity. To this end, we ought to keep the autoencoder deep and nonlinear. Unfortunately, too much nonlinearity seems to be harmful to the simulation as well. As illustrated in Fig.~\ref{fig:overfit}, with increased nonlinearity, the network (i.e., the curve in the figure) can be stretched to reach some irregular and distant poses in the training set. In the meantime, the geometry of the deformation manifold also becomes more wiggling -- same as what we experience in high-order polynomial fitting. As we know, the simulation under nonlinear reduction corresponds to travelling on the deformation manifold of $D$, driven by the generalized forces in the tangent space. A wiggling manifold could stiffen the simulation and induce artifacts.

\setlength{\columnsep}{10 pt}
There are several possible remedies of this issue. As in~\cite{fulton2019latent}, one could regularize the training data before the network training. This strategy is commonly used in training deep neural nets on a very large-scale data set that could potentially be noisy e.g., ImageNet~\cite{russakovsky2015imagenet}. However, the training data in our case are synthesized by running physical simulations, and they are \emph{noise-free}. While PCA regularization certainly prevents overfitting, it also negatively impacts the richness of the nonlinear subspace. Alternatively, CAE is also a promising method~\cite{rifai2011higher}. It injects a penalty term related to $\Vert \mathbf{J} \Vert_F$ to enhance the smoothness of $D$ so that a highly curved manifold is unlikely. Unfortunately, neither method looks attractive to us: the very reason of using DAE is to enhance the nonlinearity of the subspace, while both PCA regularization and Jacobian penalty aim to prune the subspace nonlinearity, contradicting our original motivation. The remaining option is to expand the dimension of the latent space, which is also problematic knowing that the computational cost for nonlinear model reduction is much higher than linear reduction (i.e., due to the evaluation of high-order differentiation). If the majority information encoded in the DAE is close to linear, why bother using nonlinear reduction and a deep network at the very beginning?

Our answer to this dilemma is to split the total simulation space into two \emph{orthogonal} spectra: a PCA-based linear subspace (or it could be constructed by any linear model reduction methods) $\mathcal{S}_p$ and a DAE-based nonlinear manifold $\mathcal{S}_q$ such that $\mathcal{S}_p \perp \mathcal{S}_q$. The orthogonality allows the dynamics from both subspaces to be simply super-positioned. The advantages of this subspace design are multifaceted. First of all, we can now increase the dimension of the linear subspace with a moderate cost to the overall simulation performance. Secondly, under this design, PCA basis matrix is part of the Jacobian of the overall subspace $\mathcal{S}_p \cup \mathcal{S}_q$. Therefore, the simulation does not  experience the locking artifact. Explicitly building the linear subspace also allows us to push the depth of the DAE as needed to capture nonlinear deformations and keep latent space highly compact at the same time. Lastly, we note that such multi-layer subspace construction has been successfully employed in simulation~\cite{harmon2013subspace,Zhang:CompDynamics:2020}, but first time in conjunction with a deep network.

\begin{figure}[h!]
  \centering
  \includegraphics[width=0.85\linewidth]{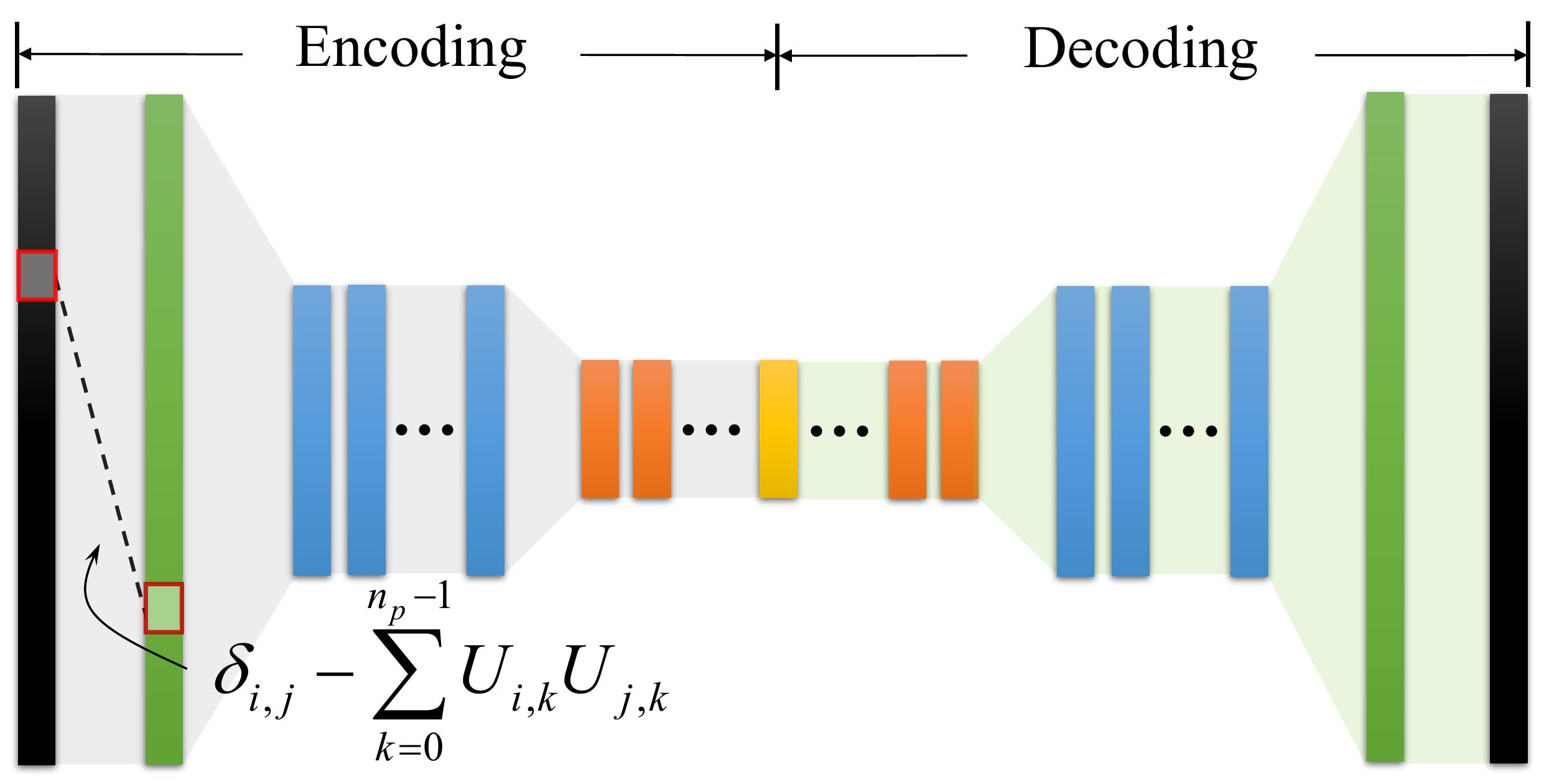}
  \caption{The network structure of our DAE. At both sides of the DAE, we append a fully connected filtering layer (in green) to remove any displacements from PCA space $\mathcal{S}_p$.}\label{fig:dae}
\end{figure}
\subsection{Network Architecture}
The network architecture of our DAE is visualized in Fig.~\ref{fig:dae}. It has a symmetric structure at encoding and decoding parts. Before training the DAE, we perform PCA over the training set to obtain basis vectors of $\mathcal{S}_p$. They are packed into the matrix $\mathbf{U}$. $\mathbf{U}$ is $N \times n_p$, where $n_p$ represents the dimensionality of $\mathcal{S}_p$. Columns in $\mathbf{U}$ are all unit vectors, and they are orthogonal to each other. To ensure $\mathcal{S}_p \perp \mathcal{S}_q$, we append a \emph{filtering} layer at both ends of the encoder and and the decoder. This filtering layer is fully connected (FC) and has fixed weights: the weight coefficient of the edge connecting $i$-th and $j$-th neurons before and after this FC layer is $\delta_{i,j} - \sum_{k=0}^{n_p-1}U_{i,k}U_{j,k}$, where $\delta_{i,j} = 1$ for $i=j$ and $0$ otherwise. In fact, this FC layer carries out a matrix-vector product of $(\mathbf{I}-\mathbf{U}\mathbf{U}^\top)\mathbf{x}$ for an input vector $\mathbf{x}$, which removes any components in $\mathbf{x}$ that generate non-zero projections in $\mathcal{S}_p$ so that the input of the encoder and the output from the decoder are all orthogonal to $\mathcal{S}_p$.

After filtering, the DAE moves to an intermediate activation part, which consists of multiple (e.g., $6$ to $8$) FC layers of the same width. The width is normally set at the order of $\log N$. Each layer is nonlinearly activated. Our activation function is quite different from other deep nets. ReLU (rectified linear unit) is a widely chosen activation, and works well in many deep learning tasks by default~\cite{nair2010rectified}. However, as ReLU is a linear activation, it fails to nonlinearly compress the training data. More importantly, ReLU is $C^0$ continuous, and a DAE only activated by ReLU may degenerate to PCA. The exponential linear unit or ELU enhances the smoothness of ReLU, but it could remain a linear activator for certain input signals. To this end, we use the \emph{trigonometric function} $\sin x$ as our activation. $\sin x$ has a derivative of an arbitrary order, and it does not have a saturated gradient at both directions. This pleasing property frees us from worrying about the vanishing gradient problem~\cite{hochreiter2001gradient} even the network is deep (over 10 layers). Finally, the feature vector is compressed to the latent space. We use $n_q$ to denote the dimension of $\mathcal{S}_{q}$. $n_q$ is a small number, typically below a couple of dozens in our experiments.

\subsection{The Simulation System}
Now, we have everything to give the formulation of the final system we need to solve. With $\mathcal{S}_p$ and $\mathcal{S}_q$ constructed, the fullspace displacement is written as:
\begin{equation}\label{eq:our_subspace}
    \mathbf{u} = \mathbf{U}\mathbf{p} + D(\mathbf{q}),\quad \text{s.t.}\quad\mathbf{U}^\top D(\mathbf{q})=0.
\end{equation}
Thanks to the orthogonality between $\mathcal{S}_p$ and $\mathcal{S}_q$, we stack Eqs.~\eqref{eq:linear_reduction}~and~\eqref{eq:nonlinear_reduction} jointly to obtain:
\begin{equation}\label{eq:our_equation_of_motion}
\widetilde{\mathbf{J}}^\top
\mathbf{M}\widetilde{\mathbf{J}}\left(\mathbf{r}-\bar{\mathbf{r}} - h \bar{\dot{\mathbf{r}}} \right)
+ \widetilde{\mathbf{J}}^\top \mathbf{f}_{fict} + h^2 \widetilde{\mathbf{J}}^\top \left(\mathbf{f}_{int} - \mathbf{f}_{ext}\right)=0,
\end{equation}
where $\mathbf{r} = [\mathbf{p}^\top, \mathbf{q}^\top]^\top$ is the generalized coordinate concatenating both $\mathbf{p}$ and $\mathbf{q}$, and $\widetilde{\mathbf{J}} = [\mathbf{U}, \mathbf{J}]\in\mathbb{R}^{N \times (n_p + n_q)}$. Here, $\mathbf{f}_{fict}$ is in the same form of Eq.~\eqref{eq:fict} because it vanishes in $\mathcal{S}_p$. Eq.~\eqref{eq:our_equation_of_motion} can then be concisely written as $\phi(\mathbf{r}) = 0$. Its Jacobian is a $(n_p + n_q) \times (n_p + n_q)$ matrix:
\begin{multline}\label{eq:system_mat}
\frac{\partial \phi}{\partial \mathbf{r}} = 
\widetilde{\mathcal{H}}^\top
\mathbf{M}\widetilde{\mathbf{J}}
\left(
\mathbf{r} - \bar{\mathbf{r}} - h \bar{\dot{\mathbf{r}}}
\right) + \widetilde{\mathcal{H}}^\top \mathbf{f}_{fict}
+
\widetilde{\mathbf{J}}^\top
\mathbf{M}
\left[\mathbf{U}, \Delta\mathbf{J} + \mathbf{J}\right] \\
- h^2 \widetilde{\mathcal{H}}^\top
(\mathbf{f}_{int} - \mathbf{f}_{ext}) + h^2 \left(\widetilde{\mathbf{J}}^\top
\frac{\partial \mathbf{f}_{int}}{\partial \mathbf{u}}
\widetilde{\mathbf{J}}\right),
\end{multline}
which needs to be updated and solved at each Newton iteration. Here, $\widetilde{\mathcal{H}} = \partial^2 \mathbf{u}/\partial \mathbf{r}^2 = \mathtt{diag}(0, \mathcal{H})$. The most involving term is $\Delta \mathbf{J}$, which is defined as:
\begin{equation}\label{eq:delta_j}
\Delta \mathbf{J} \triangleq \left(\mathbb{S}\cdot(\mathbf{q} - \bar{\mathbf{q}})\right) \cdot (\mathbf{q} - \bar{\mathbf{q}}) + \mathcal{H}\cdot(3\mathbf{q} - 3\bar{\mathbf{q}} - h \bar{\dot{\mathbf{q}}}).
\end{equation}
In order to compute $\Delta \mathbf{J}$, we need to calculate $\mathbb{S}$, an $N \times n_q \times n_q \times n_q$ forth tensor, and it is the third-order derivative of DAE: ${\partial^3 D(\mathbf{q})}/{\partial \mathbf{q}^3}$. If we choose to use first-order or quasi-Newton solvers~\cite{liu2017quasi}, the computation of $\mathbb{S}$ could be avoided, but we still need to compute the Hessian $\mathcal{H}$. Nevertheless, for subspace simulation with a small-size system matrix, Newton's method with a direct linear solver like Cholesky is always preferred.

Evaluating high-order differentiation of a deep net is not intuitive. Currently, gradient-based optimization is the mainstream solution for the network training, where the network gradient can be computed via BP. Second- and high-order derivatives are not well supported and are not efficient enough for subspace simulation tasks. Next, we discuss how we solve this technical challenge by exploiting the complex-step finite difference scheme.

\section{High-order Differentiability via CSFD}\label{sec:CSFD}
Computing the derivative of a function is omnipresent in physics-based simulation. It is typically done by inferring analytic form of the derivative function by hand or with assistance from some \emph{symbolic differentiation} software like \texttt{Mathematica}~\cite{wolfram1999mathematica}.
Alternatively, it is also possible to approximate the derivative numerically. The finite difference is the most commonly-used method, which applies a small perturbation $h$ to the function input and the first-order function derivative can be estimated as: $f'(x) \approx (f(x + h) - f(x))/{h}$. However, it is also known finite difference suffers with the numerical stability issue named \emph{subtractive cancellation}~\cite{luo2019accelerated}. This limitation could be avoided by complex-step finite difference or CSFD~\cite{martins2003complex,luo2019accelerated}.

\subsection{First- and High-order CSFD}
Let $(\cdot)^*$ denote a complex variable, and suppose $f^*:\mathbb{C}\rightarrow\mathbb{C}$ is differentiable around $x_0^*=x_0+0i$. With an imaginary perturbation $hi$, $f^*$ can be expanded as:
\begin{equation}\label{eq:complex_taylor}
f^*(x_0 + hi) = f^*(x_0) + f^{*'}(x_0)\cdot{hi} +\mathbf{O}(h^2).
\end{equation}
We can ``promote'' a real-value function $f$ to be a complex-value one $f^*$ by allowing complex inputs while retaining its original computation procedure. Under this circumstance, we have $f^*(x_0) = f(x_0), f^{*'}(x_0) = f'(x_0) \in \mathbb{R}$. Extracting imaginary parts of both sides in Eq.~\eqref{eq:complex_taylor} yields:
\begin{equation}\label{eq:im}
\mathtt{Im}\big(f^*(x_0 + hi)\big) = \mathtt{Im}\big(f^*(x_0)+f^{*'}(x_0)\cdot{hi}\big)+\mathbf{O}(h^3).
\end{equation}
Note that the error term ($\mathbf{O}(h^3)$) in Eq.~\eqref{eq:im} is cubic because the quadratic term of $h$ in Eq.~\eqref{eq:complex_taylor} is a real quantity and is excluded by $\mathtt{Im}$ operator. We then have the first-order CSFD approximation:
\begin{equation}\label{eq:cfd}
f'(x_0)=\frac{\mathtt{Im}\big(f^*(x_0+hi)\big)}{h}+\mathbf{O}(h^2)\approx\frac{\mathtt{Im}\big(f^*(x_0+hi)\big)}{h}.
\end{equation}
It is clear that Eq.~\eqref{eq:cfd} does not have a subtractive numerator, meaning it only has the round-off error regardless of the size of the perturbation $h$. If $h \sim \sqrt{\epsilon}$ i.e., around $1\times10^{-16}$, CSFD approximation error is at the order of the machine epsilon $\epsilon$. Hence, CSFD can be as accurate as analytic derivative because the analytic derivative also has a round-off error of $\epsilon$.

The generalization of CSFD to second- or even higher-order differentiation is straightforward by making the perturbation a \emph{multicomplex} quantity~\cite{lantoine2012using,nasir2013new}. The multicomplex number is defined recursively: its base cases are the real set $\mathbb{C}^0 = \mathbb{R}$, and the regular complex set $\mathbb{C}^1 = \mathbb{C}$. $\mathbb{C}^1$ extends the real set ($\mathbb{C}^0$) by adding an imaginary unit $i$ as: $\mathbb{C}^1 = \{x + yi |x,y \in \mathbb{C}^0 \}$. The multicomplex number up to an order of $n$ is defined as: $\mathbb{C}^n = \{ z_1 + z_2 i_n | z_1, z_2 \in \mathbb{C}^{n-1} \}$. Under this generalization, the multicomplex Taylor expansion becomes:
\begin{multline}\label{eq:mc_taylor}
f^{\star}(x_0+hi_1+\cdots+hi_n)=f^{\star}(x_0)+f^{\star'}(x_0) h\sum_{j=1}^ni_j \\
+\frac{f^{\star''}(x_0)}{2} h^2 \big(\sum_{j=1}^ni_j\big)^2
+\cdots \frac{f^{\star(k)}}{k!} h^k \big(\sum_{j=1}^ni_j\big)^k\cdots
\end{multline}
Here, $\left(\sum i_j\right)^k$ can be computed following the \emph{multinomial theorem}, and it contains products of mixed $k$ imaginary directions for $k$-th-order terms. For instance, the second-order CSFD formulation can then be derived as follows:
\begin{equation}\label{eq:mcfd_hessian}
\left\{
\begin{array}{l}
  \displaystyle \frac{\partial^2f(x,y)}{\partial x^2} \approx \frac{\mathtt{Im}^{(2)}\big(f(x+hi_1+hi_2, y)\big)}{h^2}\\

  \displaystyle \frac{\partial^2f(x,y)}{\partial y^2} \approx \frac{\mathtt{Im}^{(2)}\big(f(x, y+hi_1+hi_2)\big)}{h^2}\\

  \displaystyle \frac{\partial^2f(x,y)}{\partial x \partial y} \approx \frac{\mathtt{Im}^{(2)}\big(f(x+hi_1, y+hi_2)\big)}{h^2},
\end{array}\right.
\end{equation}
where $\mathtt{Im}^{(2)}$ picks the mixed imaginary direction of $i_1 i_2$. One can easily tell from Eq.~\eqref{eq:mcfd_hessian} that second-order CSFD is also subtraction-free making them as robust/accurate as the first-order case. 
With CSFD, we augment the DAE to allow each neuron to house a complex or a multicomplex quantity. Therefore, the input perturbation can be passed through the network for computing its derivative values.

\subsection{Differentiation under Tensor Contraction}
A limitation of CSFD lies in its dependency on the perturbation. If the function takes $m$ input variables e.g., an $m$-dimension vector, CSFD needs to evaluate the function for $m$ times in order to compute its first-order derivative. In our case, the function is shaped as a DAE. More precisely, $D(\mathbf{q})$ corresponds to the decoding part of the network (Fig.~\ref{fig:dae}). We need to take a forward pass of the decoding network as one function evaluation. The total number of network forwards goes up exponentially with respect to the order of differentiation. Therefore, computing $\mathbb{S}$ in Eq.~\eqref{eq:delta_j} requires $n_q^3$ network forwards per Newton iteration, which is further scaled by the complexity $D$. This is too expensive for real-time simulation even on GPU.

An important contribution of this work is to efficiently enable high-order differentiability of DAE (or other deep networks) while eliminating excessive network perturbations. Our method is based on two following key observations:
\begin{itemize}[leftmargin = 10 pt]
    \item 
    In CSFD, the imaginary parts can be somehow understood as the differential change induced by the perturbation. Under a straight usage, CSFD is analogous to forward automatic differentiation (AD)~\cite{guenter2007efficient}, but with much better generalization to higher orders. The potential of CSFD is maximized if the function has a high-dimension output so that one function evaluation gives you more information of the differentiation. Conversely, the BP procedure of a neural net is essentially a reverse AD~\cite{baydin2017automatic} -- its efficiency is optimal when the input of a network is in high dimension. This is exactly the case in neural net training, where we have a large number of network parameters as the function input. It is clear that CSFD and BP nicely complement each other so that we can choose the direction of network propagation accordingly. \\
    
    \item While high-order differentiation produces high-order tensors, those tensors are rarely needed in its original form. In most cases, they are to be ``reduced'' by tensor contractions with other tensors left and right to them. Those reduction operations allow us to apply the perturbation \emph{collectively}, not at an individual variable but in the form of vector or tensor. 
\end{itemize}

\subsection{Right Contraction via Directional Derivative}\label{subsec:right_contraction}
We now elaborate our method first with a toy example. Consider $f: \mathbb{R}^m \rightarrow \mathbb{R}$. Computing its Hessian ($\mathbf{H} = \nabla^2 f$) will need $m^2$ perturbations with second CSFD (Eq.~\eqref{eq:mcfd_hessian}). However, if $\nabla^2 f$ is also contracted with a right vector $\mathbf{a}$, $\mathbf{H}\mathbf{a}$ can actually be evaluated much more efficiently as:
\begin{equation*}
\begin{array}{ll}
   \displaystyle  & \displaystyle [\mathbf{H}(\mathbf{x})\mathbf{a}]_k = \sum_{l=0}^{m-1} \lim_{h \rightarrow 0} \frac{[\nabla f(\mathbf{x} + h \mathbf{e}_l) - \nabla f(\mathbf{x})]_k}{h} \cdot [\mathbf{a}]_l,\\
   \displaystyle \Rightarrow  & \displaystyle [\mathbf{H}(\mathbf{x})\mathbf{a}]_k = \sum_{l=0}^{m-1} \lim_{h \rightarrow 0}\frac{[\nabla f(\mathbf{x} +  [\mathbf{a}]_l h \mathbf{e}_l) - \nabla f (\mathbf{x})]_k}{h},\\
   \displaystyle \Rightarrow & \displaystyle \mathbf{H}\mathbf{a} = \lim_{h \rightarrow 0} \frac{\nabla f(\mathbf{w} + h \mathbf{a}) - \nabla f(\mathbf{x})}{h} \approx \frac{\mathtt{Im}(\nabla f(\mathbf{x} + hi \cdot \mathbf{x}))}{h}.
\end{array}
\end{equation*}
Here, $[\cdot]_k$ gives $k$-th element of vector, and $\mathbf{e}$ is the canonical bases. In the second line of the derivation, we substitute $h$ with $[\mathbf{a}]_l h$ to cancel the multiplication of $[\mathbf{a}]_l$. One may now recognize that $\mathbf{H}\mathbf{a}$ is essentially the \emph{directional derivative} of $\nabla_{\mathbf{a}}f$. 

\setlength{\columnsep}{10 pt}
\begin{wrapfigure}{r}{0.35\linewidth}
\includegraphics[width = \linewidth]{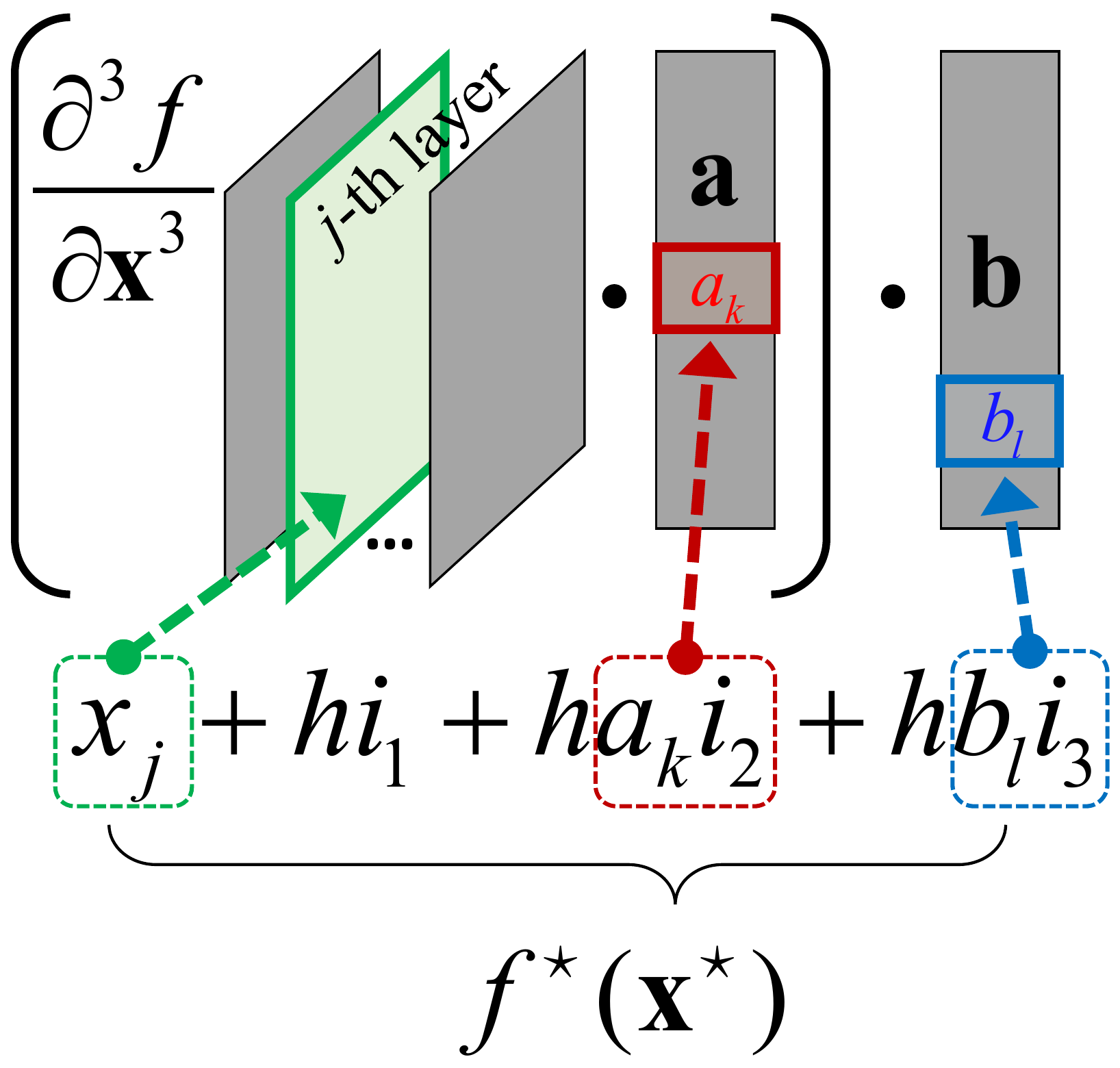}
\caption{Right contraction can be dealt with by applying CSFD perturbation collectively.}\label{fig:dd}
\vspace{-10 pt}
\end{wrapfigure}
This finding is not new and has been used in Jacobian-free solvers~\cite{knoll2004jacobian}. However, we note that this strategy can also be generalized for high-order cases. As shown in Fig.~\ref{fig:dd}, one differentiation operation lifts the order of the resulting tensor by one. A right contraction of the tensor undoes this expansion so that the perturbation can be applied together. Now let us advert to the forth tensor $\mathbb{S}$ in Eq.~\eqref{eq:delta_j}. Its exact form is of less interest to us. Instead, we would like to compute the matrix after two contractions with $\mathbf{q}-\bar{\mathbf{q}}$. To this end, we apply a collective third-order multicomplex perturbations to the decoding DAE for $n_q$ times. The $j$-th perturbation computes the $j$-th column of the resulting matrix. This perturbation is applied along the first imaginary direction $i_1$ at the $j$-th element of the DAE input $\mathbf{q}$. The perturbations in $i_2$ and $i_3$ are scaled by the corresponding elements in $\mathbf{q} - \bar{\mathbf{q}}$. Putting together, the $j$-th element, which is a third-order multicomplex quantity, of the CSFD input is:
\begin{equation}
    [\mathbf{q}^\star]_j = [\mathbf{q}]_j + hi_1 + h[\mathbf{q} - \bar{\mathbf{q}}]_j i_2 +  h[\mathbf{q} - \bar{\mathbf{q}}]_j i_3.
\end{equation}
After the forward pass, we extract the component at $i_1 i_2 i_3$ direction, and divide it by $h^3$.

\subsection{Left Contraction via Complex-step Backpropagation}\label{subsec:left_contraction}
In the simulation, there are several computations involving contraction between a left vector and a differentiation tensor such as all the $\mathcal{H}^\top$ terms in Eq.~\eqref{eq:system_mat}. In those cases, the contraction occurs at the dimension which is not expanded by the differentiation. Hence, the strategy outlined in \S~\ref{subsec:right_contraction} does not apply. Consider evaluating $\mathbf{a}\cdot\mathcal{H}$ (i.e., $\mathcal{H}^\top \mathbf{a}$). We carry out our computation with an auxiliary function $ g(\mathbf{q}) = \mathbf{a} \cdot D(\mathbf{q}) \in \mathbb{R}$. As $D$ is embodied as a neural network, this auxiliary function can also be viewed as appending an FC layer at the end of the net reducing its $N$-dimension output to a single scalar (like the loss function). Because $\mathbf{a}$ is independent on $D$, $\partial^k g/\partial \mathbf{q}^k = \mathbf{a} \cdot \partial^k D/\partial \mathbf{q}^k$. Hence, $\mathbf{a}\cdot\mathcal{H}$ can be computed as the Hessian of $g(\mathbf{q})$. Here, the reader may be reminded that $\mathcal{H}$ is a function of $\mathbf{q}$, and it is the second derivative of $d$. A standard second CSFD will need $n_p(n_p + 1)/2$ perturbations knowing $\partial^2 g/\partial \mathbf{q}^2$ is symmetric. We show that is computation can be further reduced to  $\mathbf{O}(n_q)$.

As mentioned, CSFD is most suited for differentiating functions with a high-dimension output -- $g(\mathbf{q})$ is not such a function, which outputs a single scalar. Its derivative could be more efficiently computed by reverse AD or BP. As a first-order routine however, BP only computes the gradient function $\partial g / \partial \mathbf{q}$. To this end, we inject CSFD into the BP procedure treating BP as a generic function and enabling complex arithmetic along the BP computation to perturb the gradient of $g$. It starts with a complex-perturbed forward pass of the network $g$ by adding the perturbation at one element (say the $j$-th element) of the network input as: $[\mathbf{q}^*]_j = [\mathbf{q}]_j + hi$. The feedforward of the net delivers this complex perturbation to all the neurons $[\mathbf{q}]_j$ influences. BP then ensues. During BP, all the computations are complex-based. If a neuron receives an imaginary component in the forward pass, this imaginary component participates in BP and passes complex-value feedback signals to its previous layer. After BP, all the signals at the input layer are divided by $h$ yielding one column of $\mathbf{a} \cdot \mathcal{H}$.

\begin{figure}[th!]
  \centering
  \includegraphics[width=\linewidth]{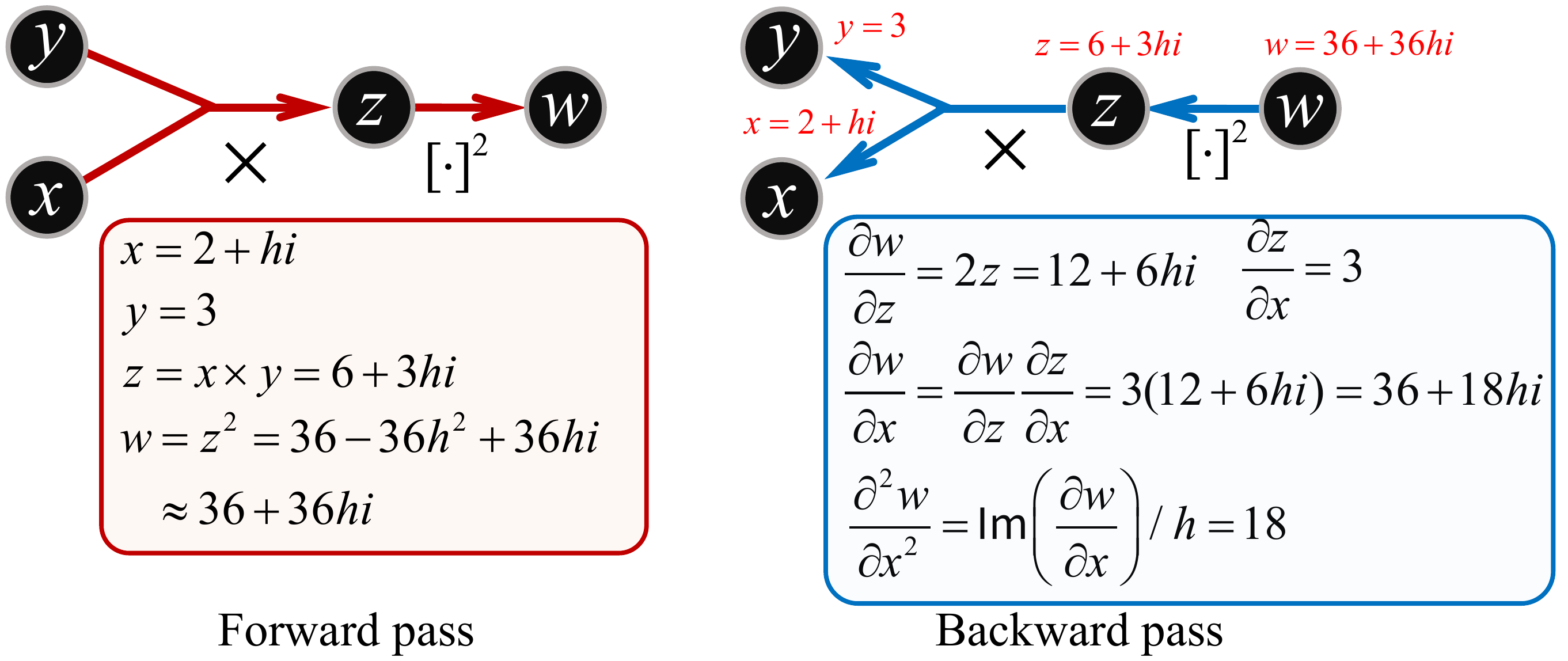}
  \caption{By augmenting BP with CSFD, we can efficiently evaluate high-order differentiation of a deep net, followed by a left-side tensor contraction.}\label{fig:bp}
\end{figure}

Fig.~\ref{fig:bp} illustrates this process with a simple net: two neurons ($x$ and $y$) multiply first, and the result ($z$) is squared to generate the output ($w$). Suppose $x = 2$ and $y = 3$, and we want to compute the second derivative of the network output with respect to $x$. The perturbation $h$ is applied to $x$ so that $x = 2+hi$, and all sequential neurons become complex-value. After the forward pass, BP invokes. Everything remains the same as the regular BP except the computation is in complex. For instance, $\partial w / \partial z = 2z$; as $z$ holds a complex value, $\partial w / \partial z = 12 + 6hi$ is also complex. Finally, after BP completes. The real part of $x$ gives the value of the first-order derivative -- the same as the original BP algorithm, and the imaginary part of $x$ after being divided by the input perturbation $h$ is the second derivative. Along this procedure, we follow the strategy in~\cite{luo2019accelerated} to avoid unneeded complex computations. For instance in Fig.~\ref{fig:bp}, high-order terms of $h$ is discarded in $w$. 

Thanks to CSFD, all the tensor-related computations can now be completed with $n_q$ network passes, either forward passes with CSFD or backward passes with CSFD-enabled BP. Those $n_q$ network passes can be executed in parallel on GPU as one single mini-batch. The remaining performance bottleneck is the subspace integration of reduced force and elastic Hessian. This computation is usually handled with the Cubature method~\cite{an2008optimizing}. In the next section, we discuss how we replace the classic Cubature sampling with a neural network based one to allow a pose-dependant subspace integration.

\begin{figure}[t!]
  \centering
  \includegraphics[width=\linewidth]{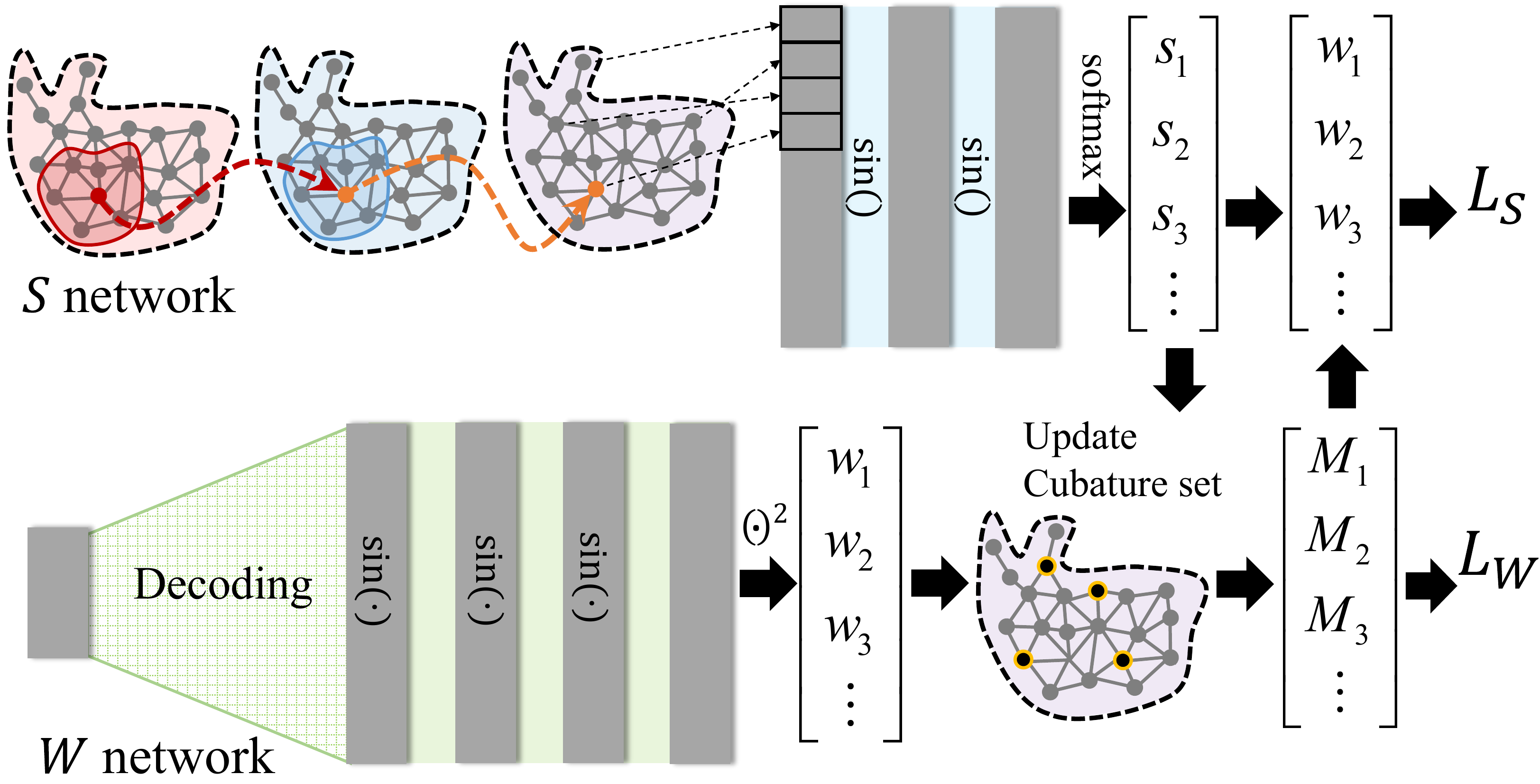}
  \caption{Neural Cubature alternates between two neural networks: $S$ and $W$. $S$ net is a GCN and selects Cubature elements with highest scores. $W$ net predicts the weight of each Cubature element. We add a square activation to ensure the non-negativeness of the output weight value. This information is then passed back to $S$ for next-round selection.}\label{fig:cubature}
\end{figure}

\section{Neural Cubature Sampling and Weighting}\label{sec:cubature}
In model reduction, it is expected that all the computations are carried out in the polynomial time of the reduced order $n_p + n_q$. For Saint Venant-Kirchhoff (StVK) material model under linear reduction, it is possible to pre-compute the polynomial coefficients for reduced force and Hessian~\cite{barbivc2005real} at the cost of $\mathbf{O}(n_p^4)$. Unfortunately, other material models do not share this convenience. A practical solution is the so-called Cubature sampling~\cite{an2008optimizing}. Cubature selects a subset of key elements (i.e., Cubature elements) such that the reduced force and reduced Hessian can be integrated only at Cubature elements with a designated non-negative weight. Cubature has been proven effective for linear model reduction. However, its na\"ive deployment for nonlinear reduction is questionable: as the tangent space varies in nonlinear cases, why should we stick with invariant Cubature weights?

Our neural Cubature consists of two networks as shown in Fig.~\ref{fig:cubature}. The first net is in charge of selecting new Cubature elements, and the second net is responsible for predicting Cubature weights. Specifically, the first neural network outputs a ``score'' for each element, and we can add \emph{multiple} elements to the Cubature set based on element's score. After updating the Cubature set $\mathcal{C}$, the second network outputs the weights of all the Cubature elements based on the input $\mathbf{r}$. Neural Cubature training alternates between those two networks. After the training, only weight prediction net participates in the simulation i.e., given a generalized coordinate $\mathbf{r}$, the neural net outputs its weights coefficients at the simulation run time, which are then used for the subspace integration.

\subsection{Cubature Selection with a GCN}
We use $S$ to denote the first neural network for Cubature element selection. $S$ is a graph convolutional network (GCN)~\cite{wu2020comprehensive}, which naturally inherits the topology of the input 3D model. The input of $S$ is the fullspace displacement $\mathbf{u}$ followed by two graph convolution layers. Each convolution layer produces eight channels. After that, another two FC layers are applied. $S$ outputs the probability $s_e$ for each element $e$ on the mesh, which is concatenated into a global probability or score vector $\mathbf{s}$. The graph convolution operation can be written as:
\begin{equation}
h_{i}^{(l+1)}=\sin\left(\sum_{j \in \mathcal{N}_{i}} \frac{1}{c_{i j}} h_{j}^{(l)} \gamma1^{(l)}\right),
\end{equation}
where $h_i^{(l)}$ represents the $i$-th vertex in the $l$-th neural network layer. $\gamma^{(l)}$ is the trainable parameter, and $\mathcal{N}$ denotes the one-ring neighborhood of $i$ on the mesh. Similar to DAE, we use $\sin(\cdot)$ for intermediate nonlinear activations. $c_{i j}=\sqrt{d_i \cdot d_j}$ is the normalization constant of edge $\langle i, j\rangle$, where $d_i$ is the degree of vertex $i$. At the last hidden layer, we use the softmax activation~\cite{goodfellow20166}, which assigns each element a probability score.

$S$ is trained in the \emph{residual space}. This scheme is inspired by the original Cubature algorithm. At the beginning, the set of Cubature elements is empty: $\mathcal{C} = \emptyset$, and the original training set consists of pose-force pairs. With some elements being selected, $\mathcal{C} \neq \emptyset$, we compute the remaining reduced force for each training data with current Cubature integration:
\begin{equation}
\overline{\mathbf{f}}(\mathbf{r}) = \widetilde{\mathbf{f}}_{int}(\mathbf{r}) - \sum M_e(\mathcal{C}, \mathbf{w}) \widetilde{\mathbf{f}}_e(\mathbf{r}).
\end{equation}
Here, $\widetilde{\mathbf{f}}_{int}(\mathbf{r})$ is the reduced internal force projected in the column space of $\widetilde{\mathbf{J}}(\mathbf{r})$. The summation iterates all the elements on the mesh. $M_e(\mathcal{C}, \mathbf{w})$ is a \emph{mask} function that removes non-Cubature weights from an input weight vector $\mathbf{w}$. In other words, $M_e(\mathcal{C}, \mathbf{w}) = [\mathbf{w}]_e$ if element $e\in\mathcal{C}$ or $0$ otherwise. Note that the dimensionality of $\mathbf{w}$ corresponds to the total number of elements on the model, and it is the output from the current weight prediction net. $\widetilde{\mathbf{f}}_e$ is the reduced force at the element $e$. Instead of adding one Cubature element each time, neural Cubature allows us to select multiple elements. After $S$ outputs scores $\mathbf{s}$, we can pick $K$ non-Cubature elements with highest scores, and update $\mathcal{C}$ accordingly. We have tested $K=5$, $K=10$, and $K=20$ and did not find much difference between them.

\subsection{Weight Prediction}
The weight prediction network $W$ takes a generalized coordinate $\mathbf{r}$ as well as the current Cubature set $\mathcal{C}$ as input, and outputs weight coefficients for all the elements $\mathbf{w}$. Specifically, $\mathbf{r}$ is first spanned to $\mathbf{u}$ with our decoder net. Network parameters at this part are fixed and do not participate in the training. Four additional FC layers with $\sin(\cdot)$ activations are followed. $W$ is not a graph network, as we believe the geometry and topology information of the model is already captured in $S$. Training the weight should be pure algebraic, and several nonlinearly activated FC layers work for this purpose well. Because $W$ is the part of the simulation (we need to obtain $\mathbf{w}$ at each time step), we also want to make sure it is light-weight and runs feedforward efficiently. Therefore, the structure of $W$ is plain and straightforward. Finally, the weight coefficients of Cubature elements should be non-negative in order to prevent extrapolation and overfitting. To this end, we put a square operation at the last layer to enforce the non-negative constraint.

The loss functions of both $S$ and $W$ resemble each other a lot:
\begin{equation}
\begin{array}{l}
\displaystyle L_S = \left\Vert \overline{\mathbf{f}}(\mathbf{r}) - \sum M_e(\mathcal{C}, \mathbf{w}) \widetilde{\mathbf{f}}_e(\mathbf{r})\right\Vert,\\
\\
\displaystyle L_W = \left\Vert \widetilde{\mathbf{f}}(\mathbf{r}) - \sum M_e(\mathcal{C}, \mathbf{w}) \widetilde{\mathbf{f}}_e(\mathbf{r}) \right\Vert.
\end{array}
\end{equation}
In practice, neural Cubature kicks off by initializing $\mathcal{C}$ as few Voronoi samples of the input model, which are passed to $W$ to start the alternating. After $\mathbf{w}$ is predicted, we feed this information to $S$ (i.e., updating the Cubature residual), which in turn, updates the Cubature set $\mathcal{C}$. Our neural Cubature is more efficient and accurate than conventional Cubature methods. Because neural Cubature picks multiple elements each time, we can also quickly build a bigger Cubature set $\mathcal{C}$.

\section{Experimental Results}\label{sec:result}
We have implemented our framework on a desktop computer with an \texttt{intel} \texttt{i7} \texttt{9700} CPU and an \texttt{nVidia} \texttt{2080} GPU. The simulation part is mostly on CPU but we move all the matrix-matrix and matrix-vector computations to GPU with \texttt{cuBLAS}~\cite{nvidia2008cublas}. The simulation is implemented with \texttt{C++}, and some linear algebra computations are based on \texttt{Eigen} library~\cite{guennebaud2010eigen}. Network training is initially carried out using \texttt{PyTorch}~\cite{paszke2019pytorch}. After we have all the network parameters, we port the resulting neural network to \texttt{CUDA}. Network BP for computing tensor contraction is also implemented with \texttt{cuBLAS}.

\subsection{Training Poses Generation}
\begin{wrapfigure}{r}{0.65\linewidth}
\vspace{-10 pt}
\includegraphics[width = \linewidth]{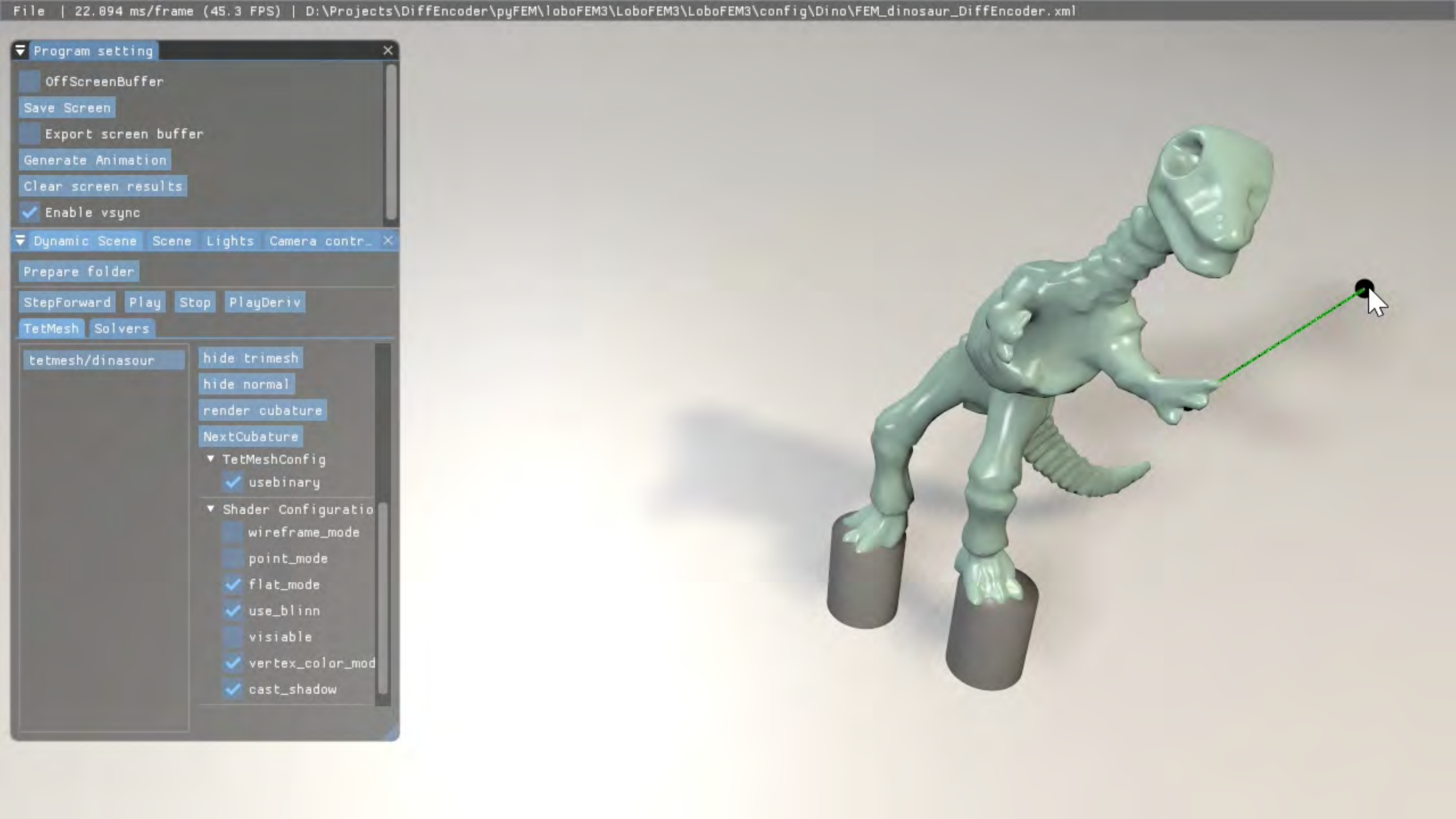}
\caption{We generate training data by applying scripted random forces to the model.}\label{fig:training}
\vspace{-10 pt}
\end{wrapfigure}

We generate training poses by running a scripted simulation. At the training stage, given a random surface vertex on the model, we select its nearby vertices within a given radius, and apply a random force to them (Fig.~\ref{fig:training}). All the simulation poses along this dynamic procedure are recorded as training data. 

\setlength{\columnsep}{10 pt}
In linear model reduction, it is common to directly sample training data within the modal space e.g., see~\cite{von2015real}. We found that this strategy is not valid for nonlinear model reduction. Here, we would like to clarify two confusing concepts: pose and basis. In linear model reduction, we care more about the basis, whose most important attribute is the direction, and its magnitude matters little. This is not the case for nonlinear reduction, where we essentially learn the underlying deformation manifold. An effective training will need samples on this manifold i.e., poses, without unnecessary scaling. Therefore, training data should be generated via real simulation.

Training poses ought not be weighted equally. In general, we prefer to better fit poses closer to the rest shape. A slightly higher fitting error may be acceptable for poses under large deformations. This is also the motivation in the linear model reduction of scaling basis vectors by their vibration frequencies~\cite{barbivc2005real,von2015real}. However, nonlinear eigenvalues of deformable poses are difficult to be estimated. We found a good metric is the elastic energy of a given pose, which nonlinearly measures how far a deformation is away from the rest configuration. As a result, we weight the loss value of each pose by the inverse of its elastic energy. As discussed, our method also builds a linear subspace (i.e., $\mathcal{S}_p$) via PCA out of the training poses. If the training data set is too big, computing a full PCA is time-consuming. We find that a good walk-around is to randomly pick poses with smallest elastic energy to form a more compact training set for PCA, and leave DAE to extract nonlinear information out of the residual pose space.

\begin{figure}[th!]
  \centering
  \includegraphics[width=\linewidth]{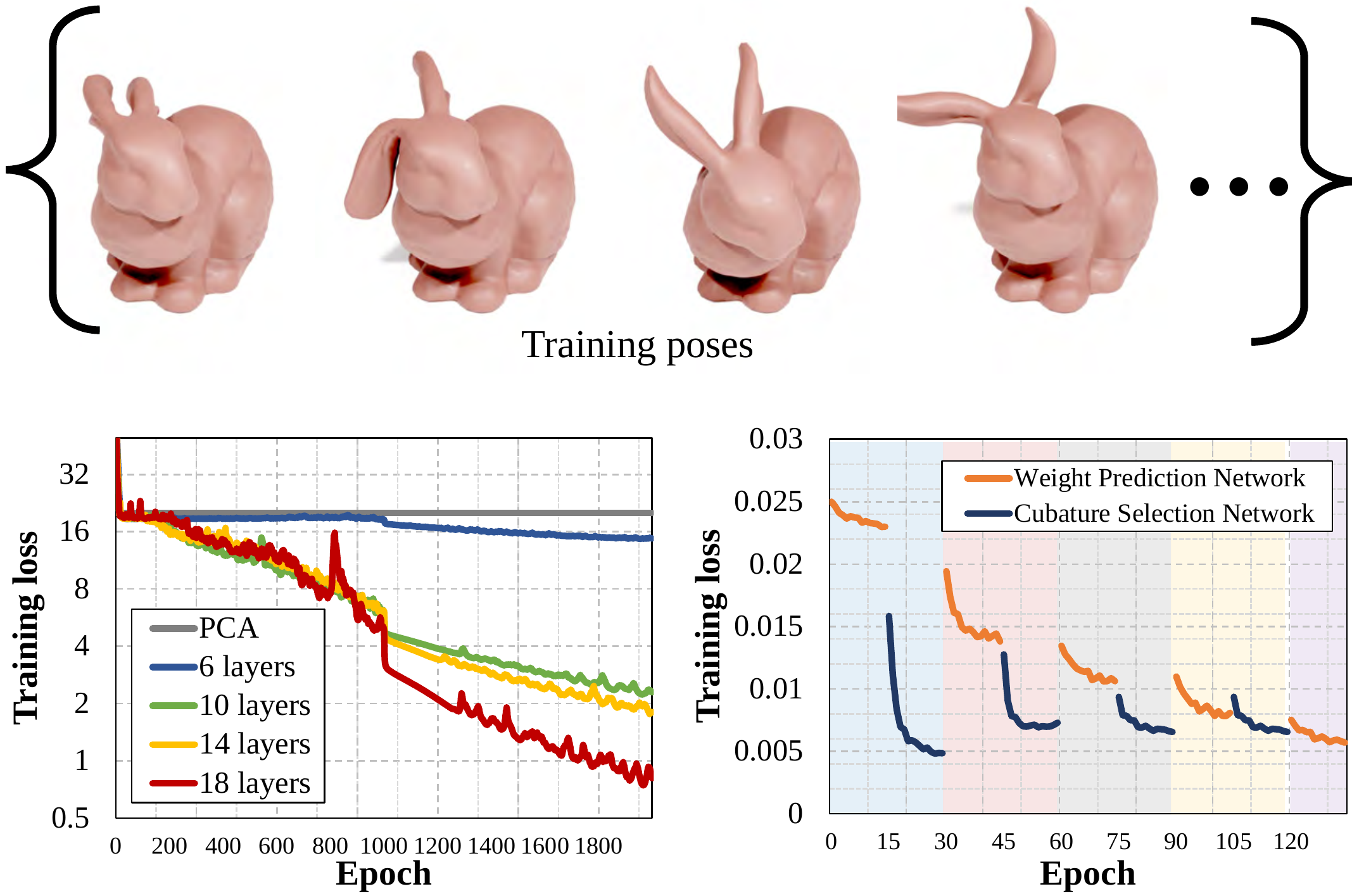}
  \caption{Network curves for training the bunny model. Neural Cubature is trained by alternating $S$ and $W$ nets, and we use the parameters from the previous alternation. Adding more layer helps reduce the training error effectively.}\label{fig:curve}
\end{figure}
\subsection{Network Training}
We use \texttt{PyTorch} and Adam for all our network training. For training DAE, we start with an initial learning rate of $0.001$. After $300$ epochs, we shrink the learning rate by $20\%$, and another $20\%$ after $3,000$ epochs. Normally, a training set includes $20,000$ poses for a model. When training the neural Cubature networks $S$ and $W$, we stick with the learning rate of $0.001$. In each alternation, we run 15 epochs for both $S$ and $W$. Depending on how many Cubature elements we want to pick, the neural Cubature training could take several thousand epochs. The total network training time is less than expected, which  takes ten to twenty minutes. Generating the training poses is the most expensive part. It often needs a couple of hours. A typical training curve is reported in Fig.~\ref{fig:curve}, which is for the bunny model. We note that the expressivity of the DAE improves with increased depth. This can be observed from Fig.~\ref{fig:curve}: if the DAE is shallow e.g., fours layers, its performance is only marginally better than PCA; but with a deeper DAE, the error decreases sharply. 


\setlength{\columnsep}{10 pt}
\begin{wrapfigure}{r}{0.55\linewidth}
\vspace{-10 pt}
\includegraphics[width = \linewidth]{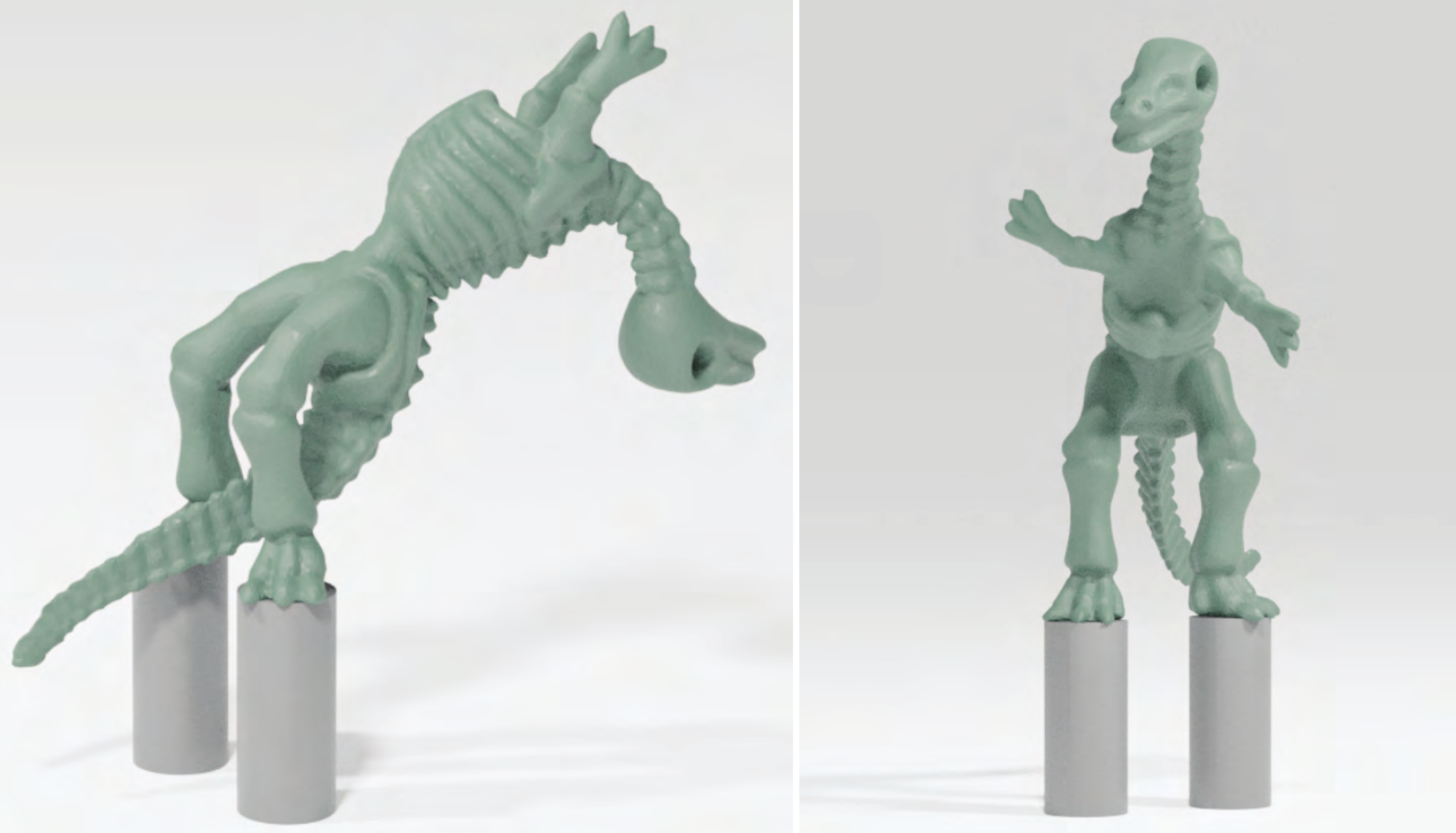}
\caption{The ground truth poses of ``bending'' (left) and ``opening arm'' (right) of the dinosaur model.}\label{fig:comp1_gt}
\vspace{-5 pt}
\end{wrapfigure}
\subsection{Comparison I: Our Method vs. Linear Model Reduction}
First, we report a comprehensive comparative experiment between our DAE-based nonlinear reduction and other commonly seen linear reduction methods including: PCA, physical modal derivative (PMD)~\cite{barbivc2005real}, and geometric modal derivative (GMD)~\cite{von2015real}. Both physical and geometric modal derivatives are based on linear modal analysis (LMA)~\cite{pentland1989good}. PMD is computed via solving a set of static equilibria around the rest shape, while GMD constructs the subspace matrix by spanning each LMA basis to nine tangent directions corresponding to its local linear transformation. In this experiment, we first compute $50$ LMA basis vectors. Based on them, we compute $50 \times (50 + 1) / 2 = 1,275$ PMD bases and $50 \times 9 = 450$ GMD bases. Finally, we apply mass-PCA as described in~\cite{barbivc2005real} to extract the subspace matrix for the linear reduction. We report the results with a dinosaur model because of its concave and non-trivial geometry.

\begin{figure*}[t!]
  \centering
  \includegraphics[width=\linewidth]{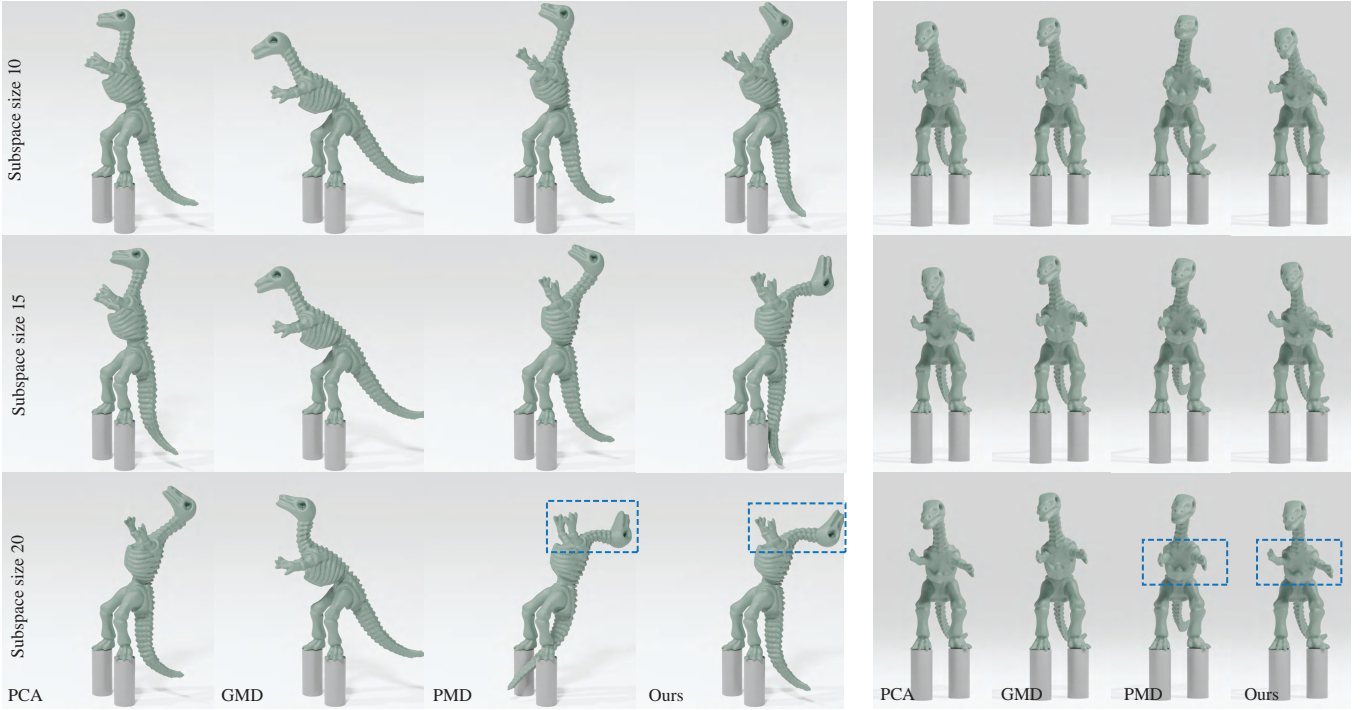}
  \caption{We compare simulation results of the dinosaur model using various model reduction methods: PCA, physical modal derivative (PMD), geometric modal derivative (GMD), and our method. The ground truth shapes are given in Fig.~\ref{fig:comp1_gt}. On the left, we globally bend the dinosaur, and on the right, we try to apply local forces at its hands.}\label{fig:comp1}
\end{figure*}

In the first set of comparison, as shown in Fig.~\ref{fig:comp1} (left), we fix the feet of the model and bend the dinosaur backwards. We compare the final poses of different reduction methods under different subspace sizes: $10$, $15$, and $20$. Our method adopts a mixed linear and nonlinear subspaces superposition, the dimensionality of each subspace is set as $n_p = 5$, $n_q = 5$; $n_p = 10$, $n_q = 5$; and $n_p = 10$, $n_q = 10$. The ground truth shape is given in Fig.~\ref{fig:comp1_gt} (left).

In this experiment, we can see a clear advantage of our method over PCA-based linear reduction. We think the reason is straightforward, DAE is known to be more expressive than conventional PCA especially for nonlinear data sets. In addition, we find that PMD also gives very good results while GMD does not perform well. We assume this is because you need to fully incorporate all $450$ geometric derivative modes in GMD to first-order approximate the derivative of LMA modes reasonably well. PMD is optimal for low-frequency deformations like this dinosaur bending. Indeed, PMD is exactly designed to capture such deformations, while our method is based on a data set generated randomly. From this perspective, it is actually encouraging to see our method yields comparable results in PMD's ``home field''. Another common trend for all the reduction methods is that the deformation improves with increased subspace dimensions.

To further verify our hypothesis, we generate another set of training poses ($500$ poses), where we only add random forces at the hands of the dinosaur. This type of deformation is local and high-frequency, which are less friendly for PMD as the bending pose. In the test, we ask the dinosaur to open its arm by applying forces to its hands outwards. All the other settings remain unchanged. As shown in Fig.~\ref{fig:comp1} (right), the difference between our method and PMD becomes more obvious in this experiment. The ground truth result is the right snapshot of Fig.~\ref{fig:comp1_gt}. Interestingly, when we narrow our training sampling at the hands, the performance of PCA also gets much better. We can see from the figure that PCA is very close to our method. This is because local deformation does not necessarily suggest higher nonlinearity. In the ``opening arm'' test, we only generate $500$ training poses, which can be fairly well captured by PCA. The advantage of nonlinear reduction is more observable when the subspace size is further condensed (e.g., when $n=10$).

\begin{figure}[th!]
  \centering
  \includegraphics[width=\linewidth]{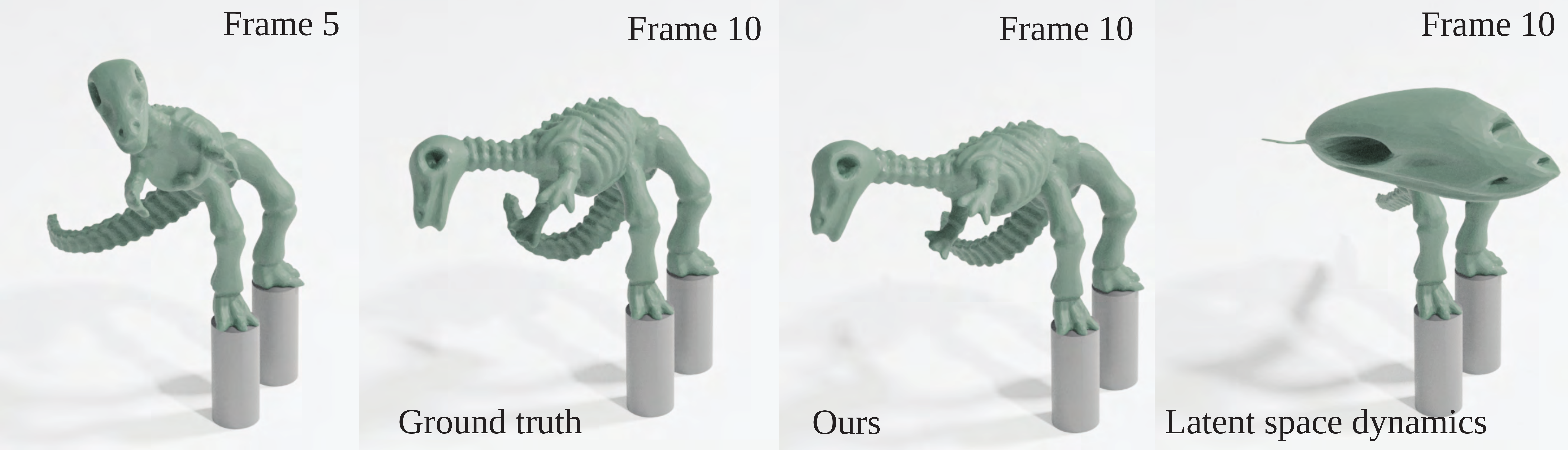}
  \caption{Latent space dynamics~\cite{fulton2019latent} uses a trimmed formulation to avoid the evaluation of high-order derivative of DAE. This simplification leads to serious artifacts when the model moves under a high velocity.}\label{fig:lsd}
\end{figure}

\subsection{Comparison II: Our Method vs. Latent Space Dynamics}
We are not the first to leverage DAE to perform nonlinear reduction. Latent space dynamics (LSD)~\cite{fulton2019latent} is closely relevant to our method. Both our method and LSD share the same high-level motivation of nonlinear subspace simulation, and both choose to use autoencoder as the machinery of the reduction in elastic simulation. Therefore, we consider LSD our major competitor. There are several key differences between our method and LSD. The most important one lies in the fact that the lack of differentiability in LSD needs a simplified formulation that ignores the fictitious force $\mathbf{f}_{fict}$ (Eq.~\eqref{eq:fict}). This could lead to significant error during the simulation when the model undergoes a high-velocity motion. 

Fig.~\ref{fig:lsd} reports snapshots of this issue. In this test, we drag the head of the dinosaur to left with an abrupt force. The artifact does not appear serious at first few frames. However, once the accumulated error reaches a certain level, the simulation diverges and cannot be recovered even we slow down the animation later. In order to have a fair comparison, we run our simulation fully in $\mathcal{S}_q$ without building the PCA space $\mathcal{S}_p$, and we do not use Cubature sampling for subspace force integration to avoid other potential error sources.

As most dynamic simulation problems are prescribed by Newton's law of motion, being able to evaluate high-order derivatives of nonlinear model reduction is a must for a successful deployment of this technique. This turns out be the key contribution of our method. In LSD, there are many smart strategies used to remedy the risk induced by the missed $\mathbf{f}_{fict}$ such as pre PCA filtering in the DAE network etc. They are all compatible with our method, but the $\mathbf{f}_{fict}$ issue does not even exist in our framework.

\begin{table}[th!]
\begin{center}
\begin{tabular}{l|c|c|c|c|c}
\whline{1.15pt}
 & $10$ & $20$ & $50$ & $100$ & $200$ \\
\whline{0.65pt}
Neural Cubature [10] & $91.1\%$ & {$62.9\%$} & {$39.1\%$} & {$23.8\%$} & {$16.3\%$} \\
Neural Cubature [5] & \blue{$87.1\%$} & \blue{$60.3\%$} & \blue{$37.1\%$} & \blue{$22.3\%$} &
\blue{$15.7\%$} \\
Greedy Cubature & {$88.4\%$} & $66.6\%$ & $47.1\%$ & $31.2\%$ & $19.2\%$  \\
\whline{1.15pt}
\end{tabular}
\end{center}
\caption{Cubature sampling error using neural Cubature and classic Cubature method. This experiment is performed on an Armadillo model with $38$K elements. Neural Cubature [10] means we add 10 elements to the Cubature set $\mathcal{C}$ based on each $S$ network prediction. Neural Cubature [5] adds 5 elements each time. Neural Cubature outperforms classic Cubature method by a substantial margin: in general neural Cubature yields $30-40\%$ less error than classic greedy Cubature method.}\label{tab:cubature}
\vspace{-20 pt}
\end{table}
\subsection{Comparison III: Neural Cubature vs. Classic Cubature}
In the next experiment, we would like to investigate the difference between our neural Cubature and the classic Cubature method. We first compare the fitting error of $10$, $20$, $50$, $100$, and $200$ Cubature elements. The results are reported in Tab.~\ref{tab:cubature}.

We can see from listed error percentages that neural Cubature outperforms classic Cubature method~\cite{an2008optimizing} in the context of nonlinear reduction. The selection network $S$ of neural Cubature uses a GCN, which captures the geometry and topology information of the input model, while classic Cubature method is solely algebraic. Another advantage of neural Cubature is its efficiency. Neural Cubature allows us to choose multiple Cubature elements each time when $S$ net predicts a score vector. We find that the Cubature training error is not sensitive to how many new Cubature elements we add to $\mathcal{C}$ every time, as long as this is a reasonable number i.e., in dozens -- picking five elements has a higher accuracy than picking ten elements, but both are better than greedy Cubature (Tab.~\ref{tab:cubature}).

\setlength{\columnsep}{10 pt}
\begin{wrapfigure}{r}{0.55\linewidth}
\vspace{-10 pt}
\includegraphics[width = \linewidth]{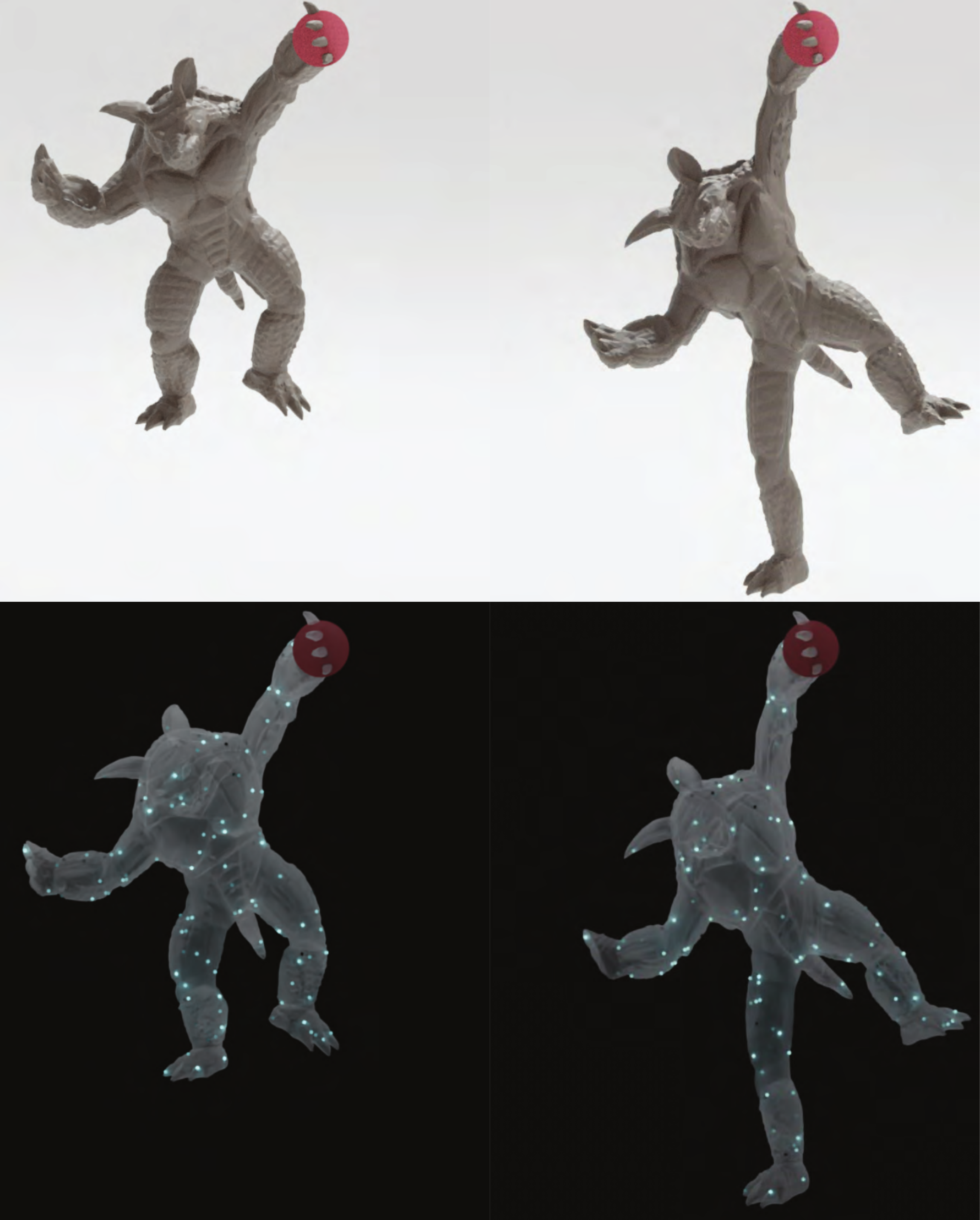}
\caption{We drag the left leg of the Armadillo. Neural Cubature with $140$ elements runs the simulation robustly. We also visualize the weight value at each Cubature element. Higher weighted elements are brighter. Greedy Cubature fails in this simulation.}\label{fig:cubature_sim}
\vspace{-5 pt}
\end{wrapfigure}
In addition, we can re-use the network parameters from the previous alternation to warm start the training for a faster convergence (e.g., see Fig.~\ref{fig:curve}). With neural Cubature, we can conveniently build a bigger $\mathcal{C}$ set. Under \texttt{CUDA}-assisted subspace integration, the neural Cubature sampling error can be effectively suppressed. This is hardly possible with classic Cubature method as we need to solve a non-negative least square problem with increasing size. Building $\mathcal{C}$ of few hundred elements would be very expensive. Fig.~\ref{fig:cubature_sim} shows a concrete experiment of simulating an Armadillo using greedy and neural Cubature strategies with $140$ Cubature elements. We fix Armadillo's hand and drag its leg downwards. Our neural Cubature with varying weights simulates this animation robustly while greedy Cubature fails ($n_p = 10$ and $n_q = 10$). In this experiment, one needs to increase greedy Cubature samples to over $250$ to reduce the error in the integration, which takes a few more training hours. 

\subsection{Comparison IV: CSFD vs. Finite Difference}
We have briefly discussed in \S~\ref{sec:CSFD} that finite difference is not numerically robust even for first-order cases. Its applicability in nonlinear reduction is unlikely possible. To verify this, we implement second- and third-order differentiation by recursively applying center finite difference. The simulation does not converge no matter how we tweak the perturbation size $h$: $h = 1E-3$, $h = 1E-5$, $h = 1E-7$, $h = 1E-9$. In fact, the simulation crashes almost immediately when finite difference is used. We believe the increased depth of the neural net imposes more challenges for finite difference to work probably. However, CSFD is robust and accurate even for high-order differentiations.


\subsection{Implementation Details}
We first use \texttt{PyTorch} to test and train our neural networks (DAE, $S$ and $W$). After the training is complete, we re-implement the net with \texttt{CUDA}, which is directly implanted in our simulation framework. Our \texttt{CUDA} port is CSFD-capable, i.e., the forward and backward pass of the network also takes multi-/complex values. This could be done by overloading the real operators with their complex or multicomplex counterparts. 

Alternatively, we choose to use the \emph{Cauchy-Riemann} (CR) formulation~\cite{ahlfors1973complex,luo2019accelerated} to achieve (multi-)complex perturbations without overloading the complex arithmetic. CR equation represents a multicomplex number in the form of a real matrix. Suppose $z^1=z^0_0+z^0_1i$, its CR form is a $2\times 2$ matrix:
\begin{equation*}
z^1=z^0_0+z^0_1i=
\left[
\begin{array}{cc}
z^0_0 & -z^0_1\\
z^0_1 & z^0_0
\end{array}
\right], \,\text{where} \;z^1\in\mathbb{C}^1\;\text{and}\;z^0_0, z^0_1\in\mathbb{C}^0=\mathbb{R}.
\end{equation*}
Here, we use the superscript $(\cdot)^n$ to denote the order of a multicomplex number. The CR matrix of $z^n$ can be constructed recursively using the CR matrices of $z^{n-1}_0$ and $z^{n-1}_1$ as:
\begin{equation}\label{eq:cr_form}
z^n=z^{n-1}_0+z^{n-1}_1i_{n}\in\mathbb{C}^n=
\left[
\begin{array}{cc}
z^{n-1}_0 & -z^{n-1}_1\\
z^{n-1}_1 & z^{n-1}_0
\end{array}
\right].
\end{equation}
Each of the $2 \times 2$ blocks in Eq.~\eqref{eq:cr_form} is a $(n-1)$-order multicomplex number, which can be further expanded with $(n-2)$-order multicomplex numbers and so on. Eventually, the CR form of $z^n$ becomes a $2^n \times 2^n$ real matrix.

With CR formula, we organize each network layer into a real layer and an imaginary layer (or multiple multicomplex layers) other than generalizing each neuron to be a complex or multiple quantity. All the computations are now in real, and we implement the forward pass of the net for FC layers with \texttt{cuBLAS}. The CR matrix multiplication is carried out block-wisely, so that we do not generate redundant multiplications corresponding to the off diagonal blocks.

The activation is on the other hand, directly implemented by launching \texttt{CUDA} threads. Fortunately, we do not have many different types of activations. Only the periodic activation function $\sin(\cdot)$ is used. To this end, we just implement its nai\"ve expression up to the third order to maximize the performance on \texttt{CUDA} without recursive variable initialization. While the expression looks verbose (e.g., in Appendix~\ref{app:sin} and the supplementary document), computing the high-order derivative of activation function only takes a small fraction of the network forwards. The major computing efforts remain at FC forward and backward passes.


\begin{table}[th!]
\begin{center}
\begin{tabular}{l|c|c|c|c|c|c}
\whline{1.15pt}
 & $\#$ Ele. & $\#$ Tri. & $n_p + n_q$  & $|\mathcal{C}|$  & $\#$ D. & FPS\\
\whline{0.65pt}
Dinosaur & $18$K & $9$K & $30+10$ & $100$ & -- & $44$ (\blue{$31\times$})\\
Armadillo & $40$K & $20$K & $30+10$ & $140$ & -- & $30$ (\blue{$35\times$}) \\
Bunny & $16$K & $8$K & $30+10$ & $100$ &  -- & $45$ (\blue{$30\times$}) \\
Cactus & $233$K & $139$K & $10 + 6$ & $6,600$ & $165$ & $2.5$ (\blue{$56\times$}) \\
Puffer ball & $625$K & $120$K & $10+5$ & $11,400$ & $321$& 1.4 (\blue{$36\times$}) \\
\whline{1.15pt}
\end{tabular}
\end{center}
\caption{Time performance of our nonlinear subspace simulator. $\#$ Ele. and $\#$ Tri. are the total numbers of elements and surface triangles on the model; $n_p+b_q$ reports the composition of our subspace configuration (per domain); $|\mathcal{C}|$ is the number of Cubature elements used; $\#$ D. is the total number of domains on the model; FPS is the simulation frame per second (and speedups compared with single-core full simulation).}\label{tab:time}
\vspace{-20 pt}
\end{table}

\begin{figure*}[t!]
  \centering
  \includegraphics[width=\linewidth]{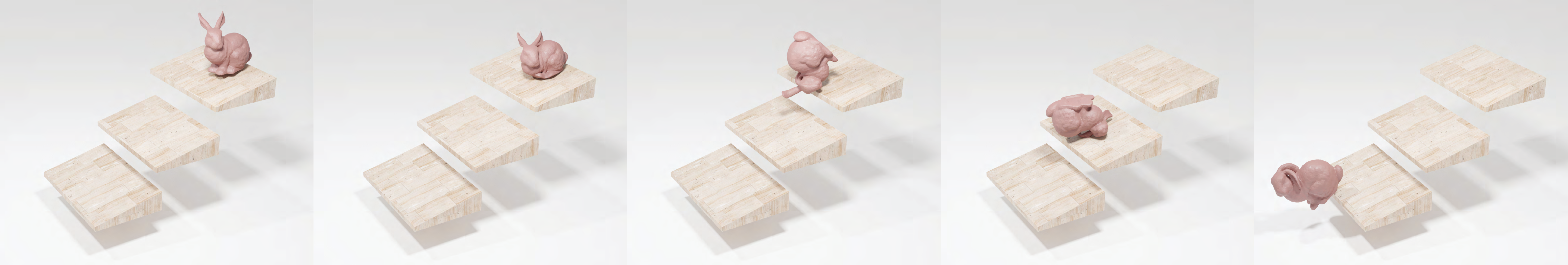}
  \caption{Falling bunny. We use generalized Newton-Euler equation to couple DAE-based model reduction with large rigid body motion to simulate free-floating objects. The subspace configuration of the bunny is $n_p =30$ and $n_q =10$.}\label{fig:bunny}
\end{figure*}

\subsection{Extensions and More Results}
Our DAE-based nonlinearly reduced simulation algorithm can be extended and integrated into other simulation frameworks at ease. For instance, we can use the generalized Newton-Euler equation to couple local deformation and rigid body dynamics~\cite{shabana2003dynamics,kim2013subspace}. The training poses need to be generated with rigid body motion removed as well in this case. Fig.~\ref{fig:bunny} reports a real-time simulation of a falling bunny on wooden stairs. We use implicit penalty force to resolve the collision and self-collision.

\begin{figure}[h!]
  \centering
  \includegraphics[width=\linewidth]{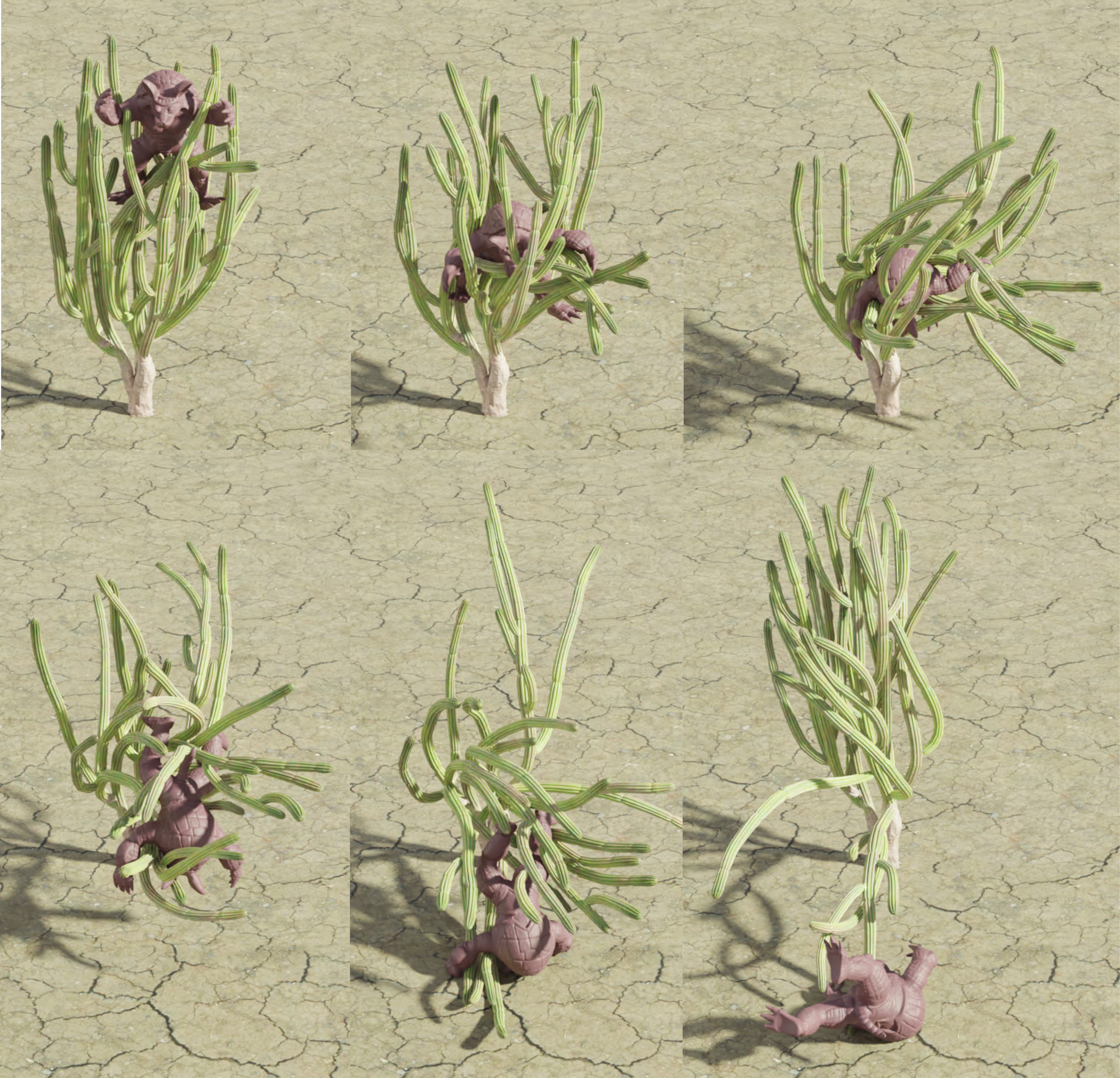}
  \caption{Dropping an Armadillo into cactus. We use deformation substructuring~\cite{barbivc2011real} method to build multi-level subspaces at the cactus. Each local domain has a compact subspace of $n_p =10$ and $n_q =6$.}\label{fig:cactus}
\end{figure}
For geometric-complex models, the advantage of nonlinear reduction could be further amplified with the domain decomposition method~\cite{barbivc2011real,yang2013boundary,wu2015unified} instead of nai\"vely increasing the global subspace size. To this end, we also couple our DAE-based nonlinear reduction with deformation substructuring~\cite{barbivc2011real}, which delivers more interesting animations with DAE-enriched local details. Two examples are reported in Figs.~\ref{fig:teaser} and \ref{fig:cactus}. The puffer ball is an ideal vehicle to deliver the advantage of this generalization. It has $320$ elastic strings with the same geometry. Therefore, the network training (for both DAE and neural Cubature) of one string can be re-used for all other strings. Thanks to its simple and symmetric geometry, depth of DAE can also be cut to $8$. The cactus example shown in Fig.~\ref{fig:cactus} is another representative case: the tree-like structure of the cactus allows an effective hierarchical deployment of nonlinear subspaces. Here, we have $165$ domains on the cactus, and each domain has a subspace of $n_p =10$ and $n_q =6$. Lastly, the simulation time performance is summarized in Tab.~\ref{tab:time}.

\section{Conclusion and Limitation}\label{sec:conclusion}
In this paper, we present a framework combining classic reduced deformable simulation with deep learning empowered data-driven approaches. We advance state-of-the-art reduction methods by plugging a deep autoencoder into the simulation pipeline. While some existing work has attempted this idea before, we are the first to address the high-order differentiability of the deep neural net in order to accurately project nonlinear dynamics of deformable solids into the tangent space of the deformation manifold. This is made possible by carefully re-engineering complex-step finite difference in the context of deep learning and complementing CSFD with reverse AD. With a CSFD-augmented BP and CSFD directional derivatives, we can evaluate the high-order derivatives of a deep net only with $\mathbf{O}(n)$ network passes. Based on this, we also propose a neural Cubature scheme that allows a more efficient Cubature sampling and more accurate weighting. Without ignoring inertia forces induced by the time-varying tangent projection, we are able to simulate deformable objects with nonlinear model reduction in real time robustly. We believe CSFD-enabled differentiability paves the way to an in-depth integration of deep neural network and physics-based simulation, which could inspire many follow-up research efforts.

There are also several limitations of our framework. First, the visual improvement of our nonlinear reduction method over existing linear reduction method is not ``wow''. After all, we are manipulating a reduced simulation with only dozens of DOFs. We believe combing neural network and other data-driven approaches used in graphics could potentially improve this issue. For instance, if the further deformation types are somehow known, we could use DAE to build a more specific nonlinear subspace as in~\cite{harmon2013subspace}. Building hierarchical DAE is also a promising solution. It may be possible to train a neural network to select multiple pre-trained DAEs to locally expand the tangent space. Model reduction is a powerful tool not only for deformable object simulation. To this end, we will further investigate how to use nonlinear reduction to improve other simulation problems like fluid, cloth, and sound synthesis.

\bibliographystyle{ACM-Reference-Format}
\bibliography{ms}

\appendix
\section{First-, Second-, and Third-order Derivative of Nonlinear Activation}\label{app:sin}
In this appendix, we give the analytic formula for first- and second-order derivative of the activation function $\sin (\cdot)$ used in our network. We give partial formulation for the third-order derivative too, which is quite verbose. For the completeness, we move the entire formulation into the supplementary document.

The first derivative of $\sin (\cdot)$ under (first-order) CSFD is:
\begin{equation*}
    \sin(a + bi_1) = \sinh(a)\cos(b) + \cosh(a)\sinh(b)i_1.
\end{equation*}

The second-order CSFD perturbation of $\sin (\cdot)$ can be written as: $ \sin(a + bi_1 + c i_2 + d i_1 i_2) = a' + b'i_1 + c' i_2 + d' i_1 i_2$, where
\begin{equation*}
\begin{aligned}
a'    &=    \sin(a)  \cosh(b)  \cosh(c)  \cos(d) -  \cos(a)  \sinh(b)  \sinh(c) \sin(d), \\ 
b'    &=    \sin(a)  \cosh(b)  \sinh(c)  \sin(d) +  \cos(a)  \sinh(b)  \cosh(c) \cos(d), \\
c'    &=    \cos(a)  \cosh(b)  \sinh(c)  \cos(d) +  \sin(a)  \sinh(b)  \cosh(c) \sin(d), \\
d'    &=    \cos(a)  \cosh(b)  \cosh(c)  \sin(d) -  \sin(a)  \sinh(b)  \sinh(c) \cos(d). \\
\end{aligned}
\end{equation*}
Similarly, we write the third-order multicomplex perturbation of $\sin (\cdot)$ as:
\begin{multline*}
    \sin(a  + b  i_1 + c  i_2 + d  i_1 i_2 + e  i_3 + f  i_1 i_3 + g  i_2 i_3 + h  i_1 i_2 i_3 ) = \\
    a' + b' i_1 + c' i_2 + d' i_1 i_2 + e' i_3 + f' i_1 i_3 + g' i_2 i_3 + h' i_1 i_2 i_3. 
\end{multline*}
The coefficient $a'$ of the real part is: 
\begin{equation}
\begin{aligned}
a' =&-\sin(a)\cosh(b)\cosh(c)\cos(d)\cosh(e)\cos(f)\cos(g)\cosh(h)\\
    &+\sin(a)\cosh(b)\cosh(c)\cos(d)\sinh(e)\sin(f)\sin(g)\sinh(h)\\
    &+\sin(a)\cosh(b)\sinh(c)\sin(d)\cosh(e)\cos(f)\sin(g)\sinh(h)\\
    &+\sin(a)\cosh(b)\sinh(c)\sin(d)\sinh(e)\sin(f)\cos(g)\cosh(h)\\
    &-\cos(a)\sinh(b)\cosh(c)\cos(d)\cosh(e)\cos(f)\sin(g)\sinh(h)\\
    &-\cos(a)\sinh(b)\cosh(c)\cos(d)\sinh(e)\sin(f)\cos(g)\cosh(h)\\
    &-\cos(a)\sinh(b)\sinh(c)\sin(d)\cosh(e)\cos(f)\cos(g)\cosh(h)\\
    &+\cos(a)\sinh(b)\sinh(c)\sin(d)\sinh(e)\sin(f)\sin(g)\sinh(h)\\
    &+\cos(a)\cosh(b)\sinh(c)\cos(d)\sinh(e)\cos(f)\sin(g)\cosh(h)\\
    &-\cos(a)\cosh(b)\sinh(c)\cos(d)\cosh(e)\sin(f)\cos(g)\sinh(h)\\
    &-\cos(a)\cosh(b)\cosh(c)\sin(d)\sinh(e)\cos(f)\cos(g)\sinh(h)\\
    &-\cos(a)\cosh(b)\cosh(c)\sin(d)\cosh(e)\sin(f)\sin(g)\cosh(h)\\
    &+\sin(a)\sinh(b)\sinh(c)\cos(d)\sinh(e)\cos(f)\cos(g)\sinh(h)\\
    &+\sin(a)\sinh(b)\sinh(c)\cos(d)\cosh(e)\sin(f)\sin(g)\cosh(h)\\
    &+\sin(a)\sinh(b)\cosh(c)\sin(d)\sinh(e)\cos(f)\sin(g)\cosh(h)\\
    &-\sin(a)\sinh(b)\cosh(c)\sin(d)\cosh(e)\sin(f)\cos(g)\sinh(h).
\end{aligned}
\end{equation}
The exact formulation of $b'$, $c'$, $d'$, $e'$, $f'$, $g'$, and $h'$ can be found in the supplementary document.

\end{document}